\documentclass[12pt, letterpaper]{article}
\usepackage{xeCJK} 
\usepackage{bm}
\usepackage{subcaption}
\usepackage{fullpage}
\usepackage{url}
\usepackage{graphicx}
\usepackage{float}
\usepackage{color}
\usepackage{rotating}
\usepackage{placeins} 
\usepackage{tabularx}
\usepackage{tikz}
\usepackage{subcaption}
\usepackage{graphicx}
\usetikzlibrary{arrows.meta}
\usepackage{chngcntr}
\usepackage{booktabs}
\usepackage{diagbox}  
\usepackage{multirow} 
\usepackage{graphicx} 
\usepackage{threeparttable}
\usepackage{longtable}
\usepackage{amsmath}
\usepackage{mathrsfs}
\usepackage{amssymb}
\usepackage[toc,page]{appendix}
\usepackage{float}
\usepackage{xcolor}
\usepackage{booktabs}
\usepackage{kbordermatrix}
\usepackage[citecolor=blue,colorlinks=true,filecolor=blue,urlcolor=blue]{hyperref}
\usepackage[stable]{footmisc}

\usepackage{caption}
\usepackage{makecell}
\usepackage{booktabs}
\usepackage{longtable}
\usepackage[round]{natbib}        
\usepackage{tikz}          
\usepackage{graphicx}      
\date{}

\usepackage[letterpaper, margin=1in, includehead]{geometry}
\usepackage{fancyhdr}

\begin{document}

\begin{center}
    \vspace*{1.5cm}

    {\LARGE \bfseries Compound Interpretation Is Based on Analogy} \\[1.2cm]

    {\large Tian Shen$^{1,2}$} \\
    {\large R. Harald Baayen$^{2}$} \\[0.3cm]

{\large $^{1}$School of Foreign Languages, Northwest University, Xi'an, China} \\
{\large $^{2}$Department of Quantitative Linguistics, University of Tübingen, Tübingen, 
Germany} \\[1.2cm]
    
    \textbf{Corresponding Author:} \\
    Tian Shen \\
    School of Foreign Languages \\
    Northwest University, Xi'an \\
    No.1 Xuefu Street, Chang'an District, Xi'an, Shaanxi, China\\
     ZIP: 710127 \\
    Email: tian.shen@uni-tuebingen.de; sweetilovefreedom@126.com \\
    Phone: +86-15349268975 \\[1.5cm]
    
\begin{flushleft}

\end{flushleft}         
        
\end{center}

\newpage

\title{Compound interpretation is based on analogy}

\maketitle
\thispagestyle{fancy}
\setcounter{page}{1}
\begin{abstract}
\noindent
How compound meanings are best predicted from constituent meanings remains a central question in computational models of lexical semantics. Comparing different computational models provides a way to evaluate alternative accounts of how semantic information is combined during compound comprehension. We propose a new model, the Compound Analogy Model (CAM), that predicts a compound's embedding by adding its constituent embeddings together with the average shift vectors of the two constituents' compound families.  The resulting model is parameter-free and exploits local analogical structure in the semantic space.  We evaluated CAM against the CAOSS model (Guenther and Marelli, 2023) on 44,731 Mandarin Chinese compounds. Computational experiments showed that CAM consistently achieved higher prediction accuracy than CAOSS on both training and held-out data, with the exception of three-character compounds, for which analogical generalization is constrained by both small constituent families and a pronounced imbalance in family size between the two constituents. The advantage of CAM remained when evaluation was based on frequency-defined train–test splits that better approximate generalization from familiar to novel compounds.  To assess the cognitive plausibility of the two models, we further examined whether model-derived semantic measures predict visual lexical decision latencies for two-character compounds.  Generalized additive models showed that predictors derived from CAM provided improved prediction for response latencies compared to predictors derived from the CAOSS model. In addition, the length (L2-norm) of the predicted compound embedding emerged as an important predictor, suggesting that embedding magnitude contributes independently to lexical processing.  These findings indicate that compound meaning is better characterized as local analogical generalization than as the application of a learned global linear transformation, and demonstrate that analogical semantic structure provides a cognitively plausible basis for compound comprehension.

\end{abstract}

\noindent{\bf keywords} \\ 
 Computational distributional semantics, Mandarin compounds processing, analogical reasoning
\section{Introduction}

\noindent Compounds are words that consist of two or more words. Compounding is a widespread word-formation process, and often one of the most productive means for creating names for novel concepts. At the same time, many compounds have meanings that are not straightforwardly predictable from the meanings of their constituents.  An \textit{apple tree}  is a tree that bears apples, but the meaning of \textit{hogwash}, `nonsense', is hard to predict from the meanings of \textit{hog} and \textit{wash}; \textit{moonshine} refers to illicit liquor rather than the light of the moon. Many studies have addressed the consequences of semantic transparency, or the lack thereof, for lexical processing and representation \citep{Zwitserlood:94,Sandra:90,Libben:Gibson:Yoon:Sandra:2003,libben2018does,Libben:2014,libben2006representation,Pham:Baayen:2015,gunther2019enter,ji2011benefits,janssen2008tale,Bien:Levelt:Baayen:2005}. \textcolor{black}{Compounds therefore provide a particularly informative test case for theories of semantic composition, because they require the cognitive system to relate constituent meanings to whole-word meanings while allowing for substantial variation in how predictable these relations are.}

Approaches in psychology to written comprehension of morphologically complex words can broad\-ly be divided into symbolic and subsymbolic theories. Within the symbolic camp, one theory argues that during reading the very first processing step is morphological decomposition \citep{Taft:2004,Rastle:Davis:New:2004,marantz2013no}. Others have argued that information about constituents only becomes available after a complex word as a whole has been recognized \citep{Giraudo:Grainger:2001}. Yet others have claimed that decompositional and whole-word driven access operate in parallel \citep{Baayen:Dijkstra:Schreuder:1997,kuperman2009reading}.  The disagreement concerns the stage at which meaning becomes relevant: is \textit{hogwash} first chunked into \textit{hog} and \textit{wash}, with subsequent processes linking these to words to the meaning `nonsense'?  Or is \textit{hogwash} immediately understood to mean `nonsense'?  Evidence for initial meaning-free morpho-orthographic segmentation \citep[e.g.][]{Rastle:Davis:New:2004,smolka2009semantics} is matched by counter-evidence pointing to confounds with degrees of semantic relatedness \citep{gonnerman2007graded,Baayen:Smolka:2020,Chuang:Kang:Luo:Baayen:2021,marelli2013meaning} and independently, evidence for an early role of whole-word knowledge in reading \citep{Schmidtke:Matsuki:Kuperman:2017,Pham:Baayen:2015} and evidence that more proficient readers scan complex words with fewer fixations \citep{Kuperman:VanDyke:2011}. Important for the present study is that all symbolic theories agree that at some point, the constituents of a compound and the compound itself are considered jointly.  The more a compound is semantically opaque, the greater the conflict between the meanings of the constituents and the meaning of the compound is that the cognitive system has to resolve.

For subsymbolic models it is less clear how the meanings of both constituents and the compound itself could be considered jointly. For the triangle model of \citet{Harm:Seidenberg:2004},  it might be possible to trace constituent effects back to hidden layers and whole-word recognition to activation at the semantic output layer.  By contrast, the Discriminative Lexicon Model \citep[DLM][]{Chuang:Kang:Luo:Baayen:2021,Heitmeier:Chuang:Baayen:2026} typically operates with networks without a hidden layer \citep[but see][for implementations using deep networks]{HeitmeierSchmidtLenschBaayen}, and directs attention to a compound's semantic embedding that is predicted from a compound's form embedding.  As shown by \citet{Bell:Baayen:2026} for English compounds, it is letter n-grams straddling the morpheme boundary (e.g., \textit{gw} in \textit{hogwash}) that are most informative for the meaning of the compound.  For compounds encountered during training, the model learns to exploit the high information load of these boundary n-grams. However, for novel compounds, the model has no prior experience with the boundary n-grams, and hence cannot predict their meanings.  In this, the model is not alone: \citet{schafer2020constituent} report for English that native speakers also struggle with making sense of novel compounds. Nevertheless, it is clear that language users (and not only linguists)  can see \textit{hog} and \textit{wash} in \textit{hogwash} \citep{Libben:2014}, and that non-specialists can be conscious of the discrepancy between the meaning of the compound and the meanings of its constituents.

It seems likely that in compound processing, different kinds of cognitive  processes are involved. One useful distinction is that between Basic Language Cognition (BLC) and the Enhanced Literate Mind (ELM) \citep{huettig2025enhanced}. BLC is the cognitive system that makes informal oral conversation possible. This system is mastered by all non-impaired speakers, irrespective of education and intelligence. It operates mostly subliminally, and is highly automatized. ELM, by contrast pertains to the cognitive skills that come with extensive training in literacy.  Whereas BLC is unproblematic for the general population, approximately one out of 10 people have some form of dyslexia, and for the rest of the population, reading skills vary widely \citep[see][for the consequences for the lexical processing of complex words]{Kuperman:VanDyke:2011}. The distinction between BLC and ELM is probably fairly fluid and not as  discrete as \citet{huettig2025enhanced} suggest.  Specifically, for highly proficient readers, automation of reading skills may lead to efficient subliminal processing similar to automated processing in BLC. The DLM \citep{Heitmeier:Chuang:Baayen:2026}, and probably also the triangle model \citep{Harm:Seidenberg:2004}, capture these BLC-like subliminal processes in proficient reading, and require complementation by mechanisms that underlie literate speakers' abilities to, e.g., provide ratings of how similar the meanings of compounds' constituents and the compound meanings themselves are.

\textcolor{black}{Regardless of whether lexical processing is assumed to proceed through symbolic or sub-symbolic mechanisms, all theories ultimately require an account of how constituent meanings give rise to compound meanings. Two broad accounts have been proposed. One explains compound meaning in terms of semantic relations between constituents, whereas the other derives compound meaning directly in a semantic vector space. Within the latter account, the present study tests whether compound meaning is better characterized as reflecting a learned global transformation or local analogical generalization.}

According to the CARIN theory of  \citet{Gagne:Shoben:97,Gagne:Spalding:Ji:2005}, readers have knowledge of the possible semantic relations \citep{lees1960grammar,Levy:1978} that may exist in a language, and that guide interpretation. For instance, a \textit{silk shirt} is a shirt \textsc{made of} silk, and a \textit{bookshelf} is a shelf \textsc{for} books.  According to \citet{Gagne:Shoben:97}, the semantic relation that is the most frequent for the first constituent guides interpretation. However, assigning a unique semantic relation to a compound has proven difficult. \cite{schmidtke2018individual} showed that, among 600 English compounds, only two received unanimous agreement on a single semantic relation from a predefined set of 16 relations, while the median number of relations selected for each compound exceeded five. These findings suggest that semantic composition cannot be fully captured by manually {\color{black} crafted} relational categories.

\citet{gunther2019enter,gunther2020semantic,gunther2023caoss} argue that compound interpretation is guided by a general function that takes the embeddings of the constituents of a compound as input and produces the embedding of the compound as output. {\color{black} The technical details of their CAOSS model} will be introduced in detail below (section \ref{sec:caoss}). The CAOSS model can be seen as a model that provides a measure of the semantic transparency of a compound: the more closely the model succeeds in approximating the true, gold-standard compound embedding, the more transparent that compound must be. The model can also be seen as providing a mechanism for obligatory decomposition models that specifies how the meanings of a compound's constituents can produce the meaning of the compound.  CAOSS can also be interpreted as providing a computationally explicit characterization of the cognitive subsystem that readers use when asked to provide semantic similarity ratings. From a linguistic perspective, CAOSS is important theoretically because it proposes there is a single semantic operation underlying compounding that is similar to the semantic operations proposed earlier for derivation \citep{marelli2015affixation}.
 
Although the hypothesis of a general semantic operation driving compound interpretation is interesting, 
\textcolor{black}{the present study explores an alternative computational principle: compound interpretation as local analogical generalization.}  Analogical reasoning has been formalized in many ways \citep[see, e.g.][]{Skousen:89,Skousen:2002,Nosofsky:1990,Daelemans:VandenBosch:2005}. The core idea is that generalization is based on majority trends in similarity neighborhoods: rather than applying a universal rule, the system exploits structure present among similar instances. For the understanding of the form and meaning of compounds, analogical approaches have proved to be useful in many respects.  CARIN  \citep{Gagne:Shoben:97,Gagne:Spalding:Ji:2005} is an analogical theory that defines the similarity neighborhood to be provided by the left constituent family of a compound, and selects the most frequent semantic relation in that family as the basis for generalization. \citet{schafer2020constituent} point out that the higher the uncertainty of a reader about the concept represented by the second constituent of an English compound, based on existing compounds with that second constituent, the greater the diversity of interpretations across readers, and the more difficult it is for these readers to imagine an interpretation.  These authors take the compounds sharing the same second constituent as providing the similarity neighborhood, and suggest that analogical generalization becomes more difficult the more diverse the senses of the second constituent are. 

With respect to compounds' forms, it has been observed that English compounds that share the same first, or second, constituent tend to be stressed in the same way   \citep{plag2010compound,bell2013informativity}. Another area where analogical generalization plays a role is the interfixes (also known as linking elements) that are found in German and Dutch compounds. Interfixes are rare in English (\textit{helm-s-man} is an example), but they occur frequently in German and Dutch.  In Dutch, \textit{schaap-herder} (`shepherd') has no interfix, \textit{schaap-s-kooi} (`sheep stable') has the \textit{-s-} interfix, and \textit{shaap-e-vlees} (`mutton') has the \textit{-e-} interfix. Both speakers and listeners have strong intuitions about which interfix is most appropriate for both existing and novel compounds, namely, the interfix that has the greatest prevalence in the left constituent family of a compound \citep{Krott:Baayen:Schreuder:2001,Krott:Schreuder:Baayen:Dressler:2007,Krott:Libben:Jarema:Dressler:Schreuder:Baayen:2004}. Here, the analogical similarity set is again the compound family of the first constituent, just as in CARIN theory.

Analogical generalization offers two advantages for understanding compound semantics.  First, analogical models can capture local generalizations \citep[cf.][]{Skousen:89} and can find `islands of reliability' \citep{Albright:Hayes:2003}. Second, because generalization is based on local similarity neighborhoods, atypical formations such as \textit{hogwash} are less likely to add noise to the system. In this study, we therefore explore whether  analogy-based generalization can explain compound interpretation.  We first report computational experiments comparing the accuracies of the CAOSS model and our new analogical model, henceforth the Compound Analogy Model (CAM), for both training data and held-out data. Subsequently, we evaluate measures derived from the two computational models against visual lexical decision latencies. The language that provides our test case is Mandarin Chinese. The next section introduces compounding in Mandarin Chinese.

\section{Compounding in Mandarin Chinese}

Mandarin Chinese is a language in which compounding plays a central role in lexical formation, accounting for approximately 70\% of existing words (Institute of Language Teaching and Research, 1986) and around 95\% of newly coined words \citep{ceccagno2016chinese}. Chinese is written with logograms (henceforth `characters') that mostly correspond to spoken syllables. The vast majority of characters can be used as words on their own, but some `cranberry' characters exist. For instance, in 尴尬 ({\em gan1ga4}, `awkward'), the first character is described as a ``meaningless bound form'' in the Pleco dictionary.  An educated speaker, after extensive training, will know approximately 3,500 characters. As in Chinese texts words are not separated by spaces, Chinese readers also develop excellent skills in distinguishing words in sequences of characters. Thus, Chinese is a prime candidate for a language in which it is impossible not to see the constituent characters of compounds.

Individual characters tend to be highly polysemous. For instance, the character 生 ({\em sheng1}) can function as a verb meaning `to become' in 生病 ({\em sheng1bing4}, `to get sick'), as an adjective meaning `raw' in 生鱼片 ({\em sheng1yu2pian4}, `sashimi'), and as a noun denoting a human referent in 学生 ({\em xue2sheng1}, `student').  Thus, the problem of constituents having multiple senses discussed by \citet{schafer2020constituent} for English is much exacerbated for Chinese (see Table~\ref{tab:examplesCompounds}). \textcolor{black}{This extensive polysemy is especially challenging for global compositional rules, but it also highlights the potential importance of local analogical structure among families of compounds.}

\begin{table}[htbp]
\centering
\caption{Examples of relatively transparent (left) and opaque (right) compounds with character 书 ({\em shu1}, `book').}
{\small
\begin{tabular}{lll|lll}
\hline
\multicolumn{3}{l}{\textbf{relatively transparent}} &
\multicolumn{3}{l}{\textbf{relatively opaque}} \\
\hline

\makecell[l]{\textbf{书}店\\店} &
\makecell[l]{shu1dian4\\dian4} &
\makecell[l]{bookshop\\inn, shop, store} &
\makecell[l]{秘\textbf{书}\\秘} &
\makecell[l]{mi4shu1\\mi4} &
\makecell[l]{secretary, assistant\\hidden, secret}
\\
\hline

\makecell[l]{读\textbf{书}\\读} &
\makecell[l]{du2shu1\\du2} &
\makecell[l]{read, study, attend school\\read, attend school} &
\makecell[l]{草\textbf{书}\\草} &
\makecell[l]{cao3shu1\\cao3} &
\makecell[l]{cursive script\\grass, straw, draft, careless, hasty}
\\
\hline

\makecell[l]{\textbf{书}架\\架} &
\makecell[l]{shu1jia4\\jia4} &
\makecell[l]{bookshelf\\support, frame, rack} &
\makecell[l]{背\textbf{书}\\背} &
\makecell[l]{bei4shu1\\bei4} &
\makecell[l]{recite a lesson for memory, endorse\\back, unlucky, recite, violate}
\\
\hline

\end{tabular}
}
\label{tab:examplesCompounds}
\end{table}

Chinese compounds exhibit diverse morphological structures \citep{ceccagno2007compound, ji2007lexical}, including modifier-head (e.g., 热心, {\em re4xin1}, `warm-hearted'), coordinate (e.g., 父母, {\em fu4mu3}, `parents'), verb-object (投资, {\em tou2zi1}, `investment'), subject-predicate (e.g., 地震, {\em di4zhen4}, `earthquake'), and verb-resultative constructions (e.g., 缩小, {\em suo1xiao3}, `shrink'). Whereas in English modifier-head compounds predominate, the position of functional heads in Chinese is remarkably free \citep{ShenBaayen2026}.

Previous computational studies addressing computationally the interpretation of Mandarin compounds include \citet{hsieh2024beyond} and \citet{wang2025composition}. The former study examined the predictive power of two measures for semantic transparency derived from a CAOSS model for visual lexical decision latencies, and reported that higher transparencies afforded shorter response latencies.  In what follows, we will report in detail how we replicated and extended their findings. The latter study took a large language model based on a  bidirectional transformer \citep{devlin2019bert} for Chinese, that was post-trained to predict 300-dimensional embeddings obtained with word2vec, building on the NOTCH model \citep{tseng2022character}.  \citet{wang2025composition} reported excellent prediction accuracy and also showed that measures based on their model out-performed the CAOSS model, as applied to Chinese by \citet{hsieh2024beyond}, for predicting lexical decision latencies. Their model, however, addresses a question that is very different from the question addressed by the CAOSS model, and that is also central to the present study.  The CAOSS model focuses on how well constituent embeddings predict compound embeddings for embeddings that all come from the same semantic space, thereby addressing the question of semantic compositionality. By contrast, the NOTCH model takes characters as input, uses a bi-directional transformer with more than 300 million parameters to obtain a 1024-dimensional vector that is subsequently mapped onto a 300-dimensional word2vec vector that itself was trained on different textual data.  As a consequence, the NOTCH model is not optimal for understanding the human cognition of compound interpretation. In what follows, we therefore do not follow up on this line of research,  and instead consider what can be achieved with simpler models working with straightforward (non-contextualized) embeddings that all belong to the same semantic space.

\section{The models}

This section first introduces the CAOSS model. It  then proceeds to outline our alternative model for compound semantics, the ``Compound Analogy Model''  (CAM).

\subsection{The CAOSS model}
\label{sec:caoss}

\noindent
The CAOSS model (Compounding as Abstract Operations in Semantic Space; \citealp{gunther2019enter, gunther2020trying, gunther2023caoss}) is a data-driven compositional model that represents compound meanings as linear combination of the embeddings of their constituents. For a two-constituent compound, the model assumes that its meaning can be computed as a weighted sum of the meanings of its constituents, using least squares multivariate multiple regression \citep{marelli2017compounding}. 
Formally, the CAOSS model works with three matrices with embeddings as row vectors:
\begin{itemize}
    \item $\bm{S}$: an $n \times k$ matrix of $k$-dimensional embeddings for $n$ compounds;
    \item $\bm{S}_1$: an $n \times k$ matrix with the embeddings of the $n$ compounds' first constituents;
    \item $\bm{S}_2$: an $n \times k$ matrix of the embeddings of the $n$ compounds' second constituents.
\end{itemize}

\noindent
The CAOSS model concatenates the embeddings of the two constituents, resulting in a $n \times 2k$ matrix, and uses a $2k \times k$ transformation matrix $\bm{W}$ to map the concatenated constituent embeddings onto the $k$-dimensional compound embeddings: 
\begin{equation}
[\bm{S}_1 \bm{S}_2] \bm{W} = \bm{S}. 
\end{equation}
Solving for $\bm{W}$, the compound embeddings predicted by this model, $\hat{\bm{S}}$ are given by  
\begin{equation}
\label{equo: CAOSS S}
    \hat{\bm{S}} = [\bm{S}_1\bm{S}_2] \bm{W}.
\end{equation}
\noindent
Let $\bm{W}_1$ and $\bm{W}_2$ be the $k \times k$ sub-matrices of $\bm{W}$ such that 
$$
\bm{W} = \left[\begin{array}{c}
    \bm{W}_1   \\
    \bm{W}_2 
\end{array}\right].
$$
The meanings of the first constituents in isolation are changed by the matrix $\bm{W}_1$ as follows:
\begin{equation}
\label{equo:CAOSS S1}
    \hat{\bm{S}}_1=\bm{S}_1\bm{W}_1,
\end{equation}
and likewise the meanings of the second constituents in isolation are transformed with $\bm{W}_2$ as follows: 
\begin{equation}
\label{equo:CAOSS S2}
    \hat{\bm{S}}_2=\bm{S}_2\bm{W}_2.
\end{equation}
We will refer to these compound-specific constituent meanings as the CAOSS-transcended meanings of these constituents, following \citet{libben2022lexicon}.

We illustrate this model with a toy example.  Figure~\ref{eq:CAOSS 学生} demonstrates how the model predicts the embedding for the two-character compound 学生 ({\em xue2sheng1}, `student') using randomly generated 5-dimensional embeddings. The model takes the embedding of the left character $\mathbf{s}_{1\text{学}}$ (blue row vector) and that of the right character $\mathbf{s}_{2\text{生}}$ (red row vector), then multiplies  by $\bm{W}_1$ and $\bm{W}_2$ respectively, producing two transcended vectors:
\begin{equation}
\hat{\mathbf{s}}_{1\text{学}}
=
\mathbf{s}_{1\text{学}}\bm{W}_1\label{eq:caoss_transc1}
\end{equation}
(`transcended 学') and
\begin{equation}
\hat{\mathbf{s}}_{2\text{生}}
=
\mathbf{s}_{2\text{生}}\bm{W}_2\label{eq:caoss_transc2}
\end{equation}
(`transcended 生'). To illustrate the matrix multiplication involved, the figure highlights one specific dimension. For \textbf{学}, the blue-highlighted values in $\bm{W}_1$ are multiplied by the corresponding embedding values and summed, yielding the blue-highlighted value (3.26) in the transcended 学 vector. The same process applies to \textbf{生}, producing the red-highlighted value (3.67) in the transcended 生 vector. The predicted compound embedding is then obtained by element-wise addition: $\hat{\bm{s}} = \hat{\bm{s}}_1 + \hat{\bm{s}}_2$, with the two highlighted values summing to the bolded entry (6.93) in the final predicted vector for 学生.

\begin{sidewaysfigure}
\tiny  
\parbox{\textheight}{
  \begin{eqnarray*}
    \underbrace{
    \left(
    \begin{array}{rrrrr}
      {\color{blue} 0.07} & {\color{blue} 1.24} & {\color{blue}1.39} & {\color{blue}1.83} & {\color{blue} -0.66} \\
    \end{array}
    \right)
  }_{\text{\normalsize $\mathbf{s}_{1\text{学}}$}}
    \cdot
    \underbrace{
    \left(
    \begin{array}{rrrrr}
      -1.35 &  1.09  &  {\color{blue} 0.09} &   0.37 &  -0.20 \\
      0.37  & -0.49  &  {\color{blue} 1.07} &  -2.10 &  -1.49 \\
      -2.49 &  1.11  &  {\color{blue} 0.99} &   1.38 &  -1.44 \\
      -0.67 &  0.14  &  {\color{blue} 0.26} &  -0.11 &  -2.28 \\
      0.12  & -0.83  & {\color{blue} -0.11} &   0.54 &  -0.94 \\ 
    \end{array}
    \right)}_{\mbox{\normalsize {left constituent} 
    matrix $\bm{W}_1$}} & + &
    \underbrace{
    \left(
    \begin{array}{rrrrr}
      {\color{red} 0.82} & {\color{red} 0.88} & {\color{red} -0.43} & {\color{red} -1.95} & {\color{red} 0.14} \\
    \end{array}
    \right)
    }_{\text{\normalsize $\mathbf{s}_{2\text{生}}$}}
    \cdot
    \underbrace{
    \left(
    \begin{array}{rrrrr}
      -1.12 &  -1.12 &   {\color{red} 1.17} &   0.83 &   1.06 \\
      -0.94 &   0.96 &   {\color{red} 0.69} &   0.14 &  -0.14 \\
       0.04 &   1.22 &  {\color{red} -0.66} &   1.09 &   0.52 \\
      -0.20 &   0.03 &  {\color{red} -0.95} &  -2.68 &  -1.24 \\
      -0.32 &  -1.47 &  {\color{red} -0.25} &   0.94 &   0.70 \\
    \end{array}
    \right)}_{\mbox{\normalsize {right constituent} 
     matrix $\bm{W}_2$}} \\
      & & \\
      & & \\
    \underbrace{
    \left(
     \begin{array}{rrrrr}
       -4.4 & 1.82 & {\color{blue} 3.26} & -1.22 & -7.42 \\
     \end{array}
    \right)
    }_{\text{\normalsize transcended vector $\hat{\mathbf{s}}_{1\text{学}}$}}
     & + &  
    \underbrace{
    \left(
     \begin{array}{rrrrr}
       -1.42 & -0.86 & {\color{red} 3.67} &  5.69&  3.04  \\
     \end{array} 
    \right)}_{\text{\normalsize transcended vector $\hat{\mathbf{s}}_{2\text{生}}$}}  = \\
      & & \\
      & & \\
    \underbrace{
    \left(
     \begin{array}{rrrrr}
       -5.82 & 0.95&  \mathbf{6.93} &  4.48&  -4.38 \\
     \end{array}
    \right)}_{\text{\normalsize $\hat{\mathbf{s}}_{\text{学生}}$}} & & \\
  \end{eqnarray*}
}

\caption{
The CAOSS model illustrated for predicting the  embedding of 学生 ({\em xue2sheng1}, `student'). The colour-highlighted values in the transcended embeddings are obtained by pairwise multiplication, followed by summation, of the values of the constituent embeddings and the highlighted column values in the transformation matrices. The bolded value of the predicted compound embedding is obtained by summing the two colour-highlighted values in the transcended embeddings. For simplicity and visual clarity, we illustrate the mechanism using randomly generated toy embeddings (rounded to two decimals).
}

\label{eq:CAOSS 学生}
\end{sidewaysfigure}


\subsection{The Compound Analogy Model (CAM)}

Previous studies addressing the distributional semantics of derived words reported that for transparent affixes, the embeddings of words with these affixes cluster together in semantic space. Furthermore, the vectors from the base word to the corresponding derived word, henceforth shift vectors also cluster in the embedding space.  This suggests that the meaning of a derived word can be represented as the sum of the vector of its base and its average shift vector, see, e.g., \citet{drozd2016word}. However, more refined  conditional shift vectors may be required, see \citet{shafaei2024semantic} and \citet{Chuang:Brown:2022}. 

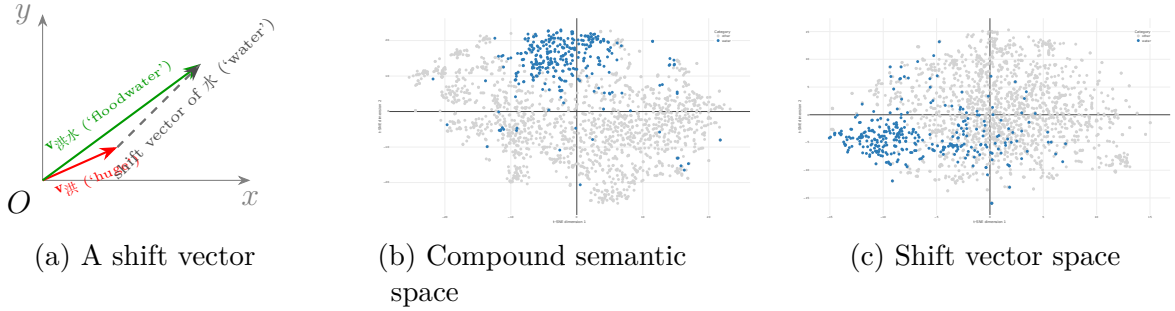
\begin{figure}[htbp]
\centering
\makebox[\textwidth][c]{%
\begin{minipage}[t]{0.3\textwidth}
\centering
\begin{tikzpicture}[
    >=Stealth,
    scale=0.55,
]
\draw[->, thin, black!50] (0,0) -- (5,0) node[below] {$x$};
\draw[->, thin, black!50] (0,0) -- (0,4) node[left]  {$y$};
\node[below left] at (0,0) {$O$};

\coordinate (Hong)     at (1.8,0.8);
\coordinate (Hongshui) at (3.8,2.8);

\draw[->, thick, red] (0,0) -- (Hong) 
    node[pos=0.65, below, sloped, font=\tiny] {$\mathbf{v}_{\text{洪 (`huge')}}$};
\draw[->, thick, green!60!black] (0,0) -- (Hongshui) 
    node[midway, above, sloped, font=\tiny] {$\mathbf{v}_{\text{洪水 (`floodwater')}}$};
\draw[->, thick, dashed, gray!60!black] (Hong) -- (Hongshui) 
    node[pos=0.75, below, sloped, font=\tiny] {shift vector of \text{水} (`water')};

\draw[gray, thin, dashed] (Hong) -- ++(2.0,2.0) coordinate (tmp);
\draw[gray, thin, dashed] (Hongshui) -- (tmp);

\end{tikzpicture}
\subcaption{A shift vector}
\end{minipage}
\hspace{0.02\textwidth}
\begin{minipage}[t]{0.3\textwidth}
\centering
\includegraphics[width=\linewidth]{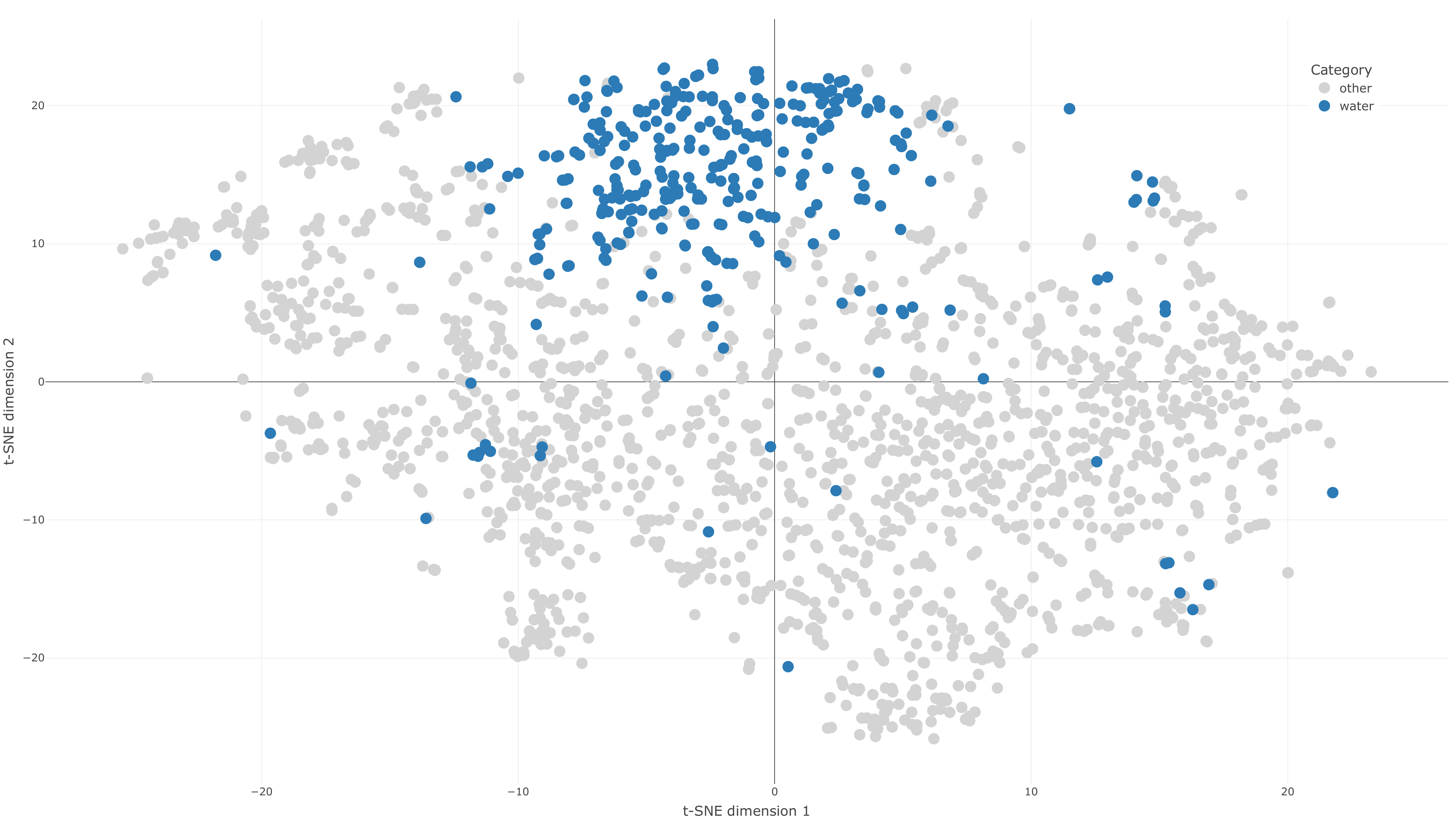} 
\subcaption{Compound semantic space}
\end{minipage}
\hspace{0.02\textwidth}
\begin{minipage}[t]{0.3\textwidth}
\centering
\includegraphics[width=\linewidth]{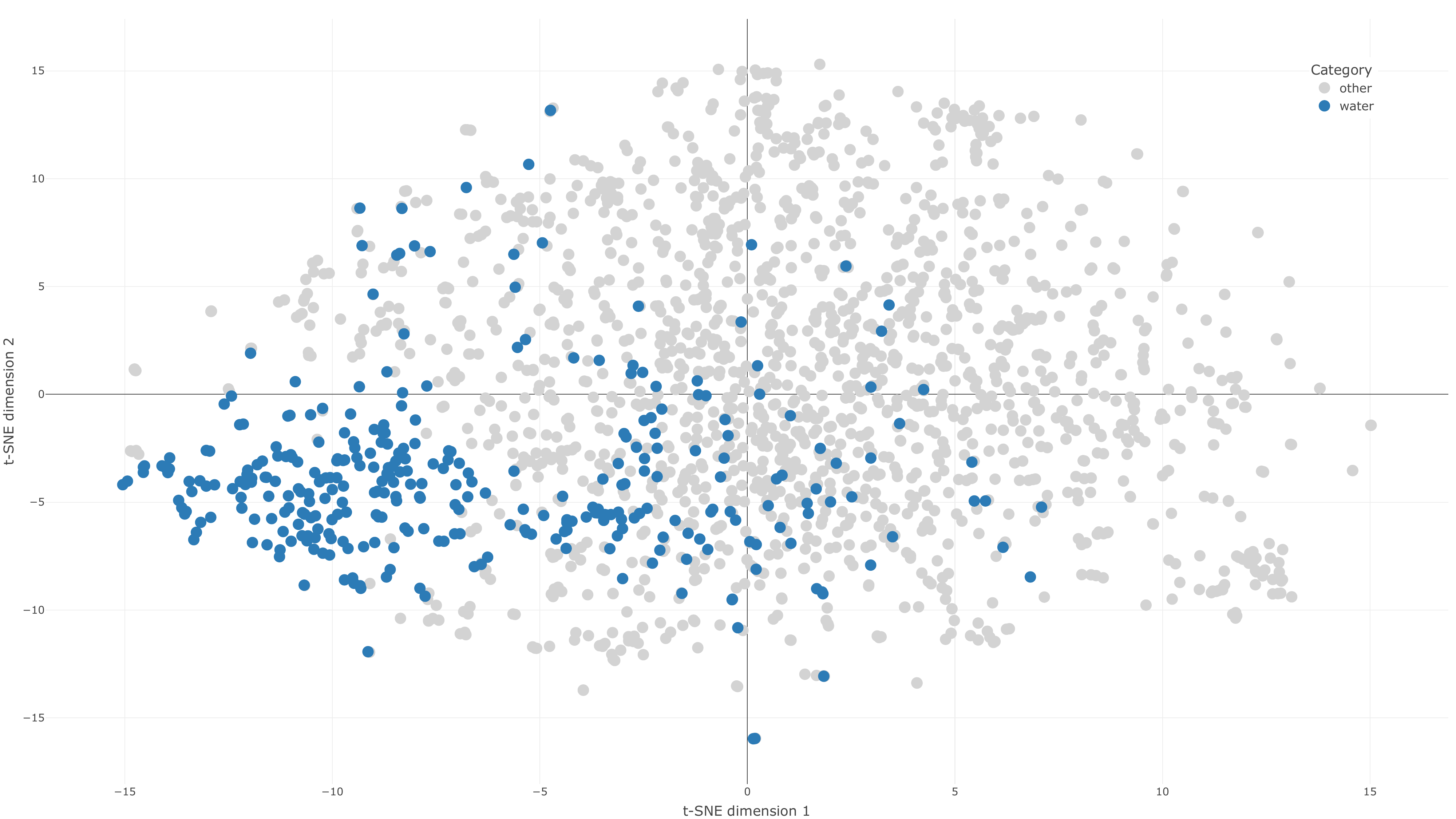} 
\subcaption{Shift vector space}
\end{minipage}
}
\caption{Core concepts of CAM. (a) A shift vector for a given constituent, here 水 ({\em shui3}, `water'), is obtained by subtracting the embedding of the other constituent, here  洪 ({\em hong2}, `huge'), from the compound embedding, here 洪水 ({\em hong2shui3}, `flood water'). (b) Compounds with 水 (`water', in blue) cluster (using t-SNE) in the embedding space. The points in grey represent compounds with nine other high-frequency monomorphemic words: 火 (`fire'), 金 (`gold'), 木 (`wood'), 土 (`earth'), 天 (`sky'), 道 (`road'), 心 (`heart'), 人 (`man'), and 名 (`name'). (c) The corresponding shift vectors of 水 likewise cluster in the corresponding shift space.}
\label{fig:shift_guide}
\end{figure}

The Compound Analogy Model (CAM) proposes that compound semantics operate on the similar additive principles. This model starts out from the observation that Mandarin compound families sharing a constituent form clusters in both the semantic space of compound embeddings and the corresponding shift vector space \citep{ShenBaayen2026}.  One such cluster is illustrated in Figure~\ref{fig:shift_guide}. The center panel illustrates the clustering in a t-SNE map of the embeddings of two-character compounds containing 水 ({\em shui3}, `water', in blue) together with compounds with other high-frequency mono-morphemic words (shown in grey).  The left panel illustrates a shift vector, and the right panel presents the shift vectors for 水. The shift vectors of 水 also show considerable clustering. The existence of these clusters suggests that in Mandarin, the semantics of compound constituent families are remarkably consistent and coherent, in ways similar to what has previously been observed for semantically transparent derivational and inflectional affixes \citep{stupak2022inquiry,nikolaev2022generating,Chuang:Brown:2022}.  Importantly, the centroid of the cluster of compound embeddings with a given compound constituent (水 in Figure \ref{fig:shift_guide}) represents the prototypical meaning of compounds with this constituent. Furthermore, the centroids of the shift vectors of given compound constituents, e.g., 水, represent the prototypical change in meaning that such constituents contribute to the compounds in which they occur.

The CAM model proposes that the meaning of a compound can be predicted from the embeddings of its constituents and the centroids of the shift vectors of these constituents.  We illustrate the model for the compound 学生 ({\em xue2sheng1}, `student'). In this example, we assume that the left and right constituent families have exactly 5 members, with 6-dimensional embeddings. (The actual constituent families are much larger, 50 and 122, and the actual embeddings are 200-dimensional.) Figure~\ref{fig:cam_xue} illustrates how the shift vector for the left constituent is calculated. For each of the compounds sharing the first constituent 水, its left-constituent family, the embeddings of these family members are brought together in a matrix. From these embeddings, we subtract the embeddings of the corresponding right constituents, resulting in a new matrix with difference vectors.  Taking the average column-wise results in the average shift vector, the centroid of the difference vectors.   Figure~\ref{fig:cam_sheng} illustrates the same procedure for the second constituent, 生.

\begin{sidewaysfigure}
\renewcommand{\kbldelim}{(}
\renewcommand{\kbrdelim}{)}
\footnotesize
\begin{eqnarray*}
{\bm{D}}_{\text{学}, 1}
&=&
\underbrace{
\kbordermatrix{
      & dim1 & dim2 & dim3 & dim4 & dim5 & dim6 \\
\text{学生} & -0.219 & \hspace*{0.8em}0.039 & -0.051 & \hspace*{0.8em}0.148 & \hspace*{0.8em}0.078 & \hspace*{0.8em}0.307 \\
\text{学术} & -0.030 & \hspace*{0.8em}0.052 & -0.264 & -0.179 & \hspace*{0.8em}0.036 & -0.541 \\
\text{学者} & \hspace*{0.8em}0.080 & \hspace*{0.8em}0.037 & -0.240 & -0.179 & -0.015 & -0.198 \\
\text{学费} & -0.466 & -0.276 & -0.125 & -0.067 & \hspace*{0.8em}0.121 & -0.134 \\
\text{学问} & -0.074 & \hspace*{0.8em}0.023 & -0.318 & \hspace*{0.8em}0.089 & \hspace*{0.8em}0.220 & -0.030 \\
}}_{\text{Compound embeddings } }
-
\underbrace{
\kbordermatrix{
      & dim1 & dim2 & dim3 & dim4 & dim5 & dim6 \\
\text{生} & 0.187 & -0.149 & -0.102 & -0.073 & 0.202 & -0.081 \\
\text{术} & -0.124 & 0.175 & -0.081 & -0.169 & 0.356 & 0.001 \\
\text{者} & 0.158 & 0.038 & -0.089 & -0.108 & 0.211 & -0.210 \\
\text{费} & -0.073 & -0.090 & 0.141 & -0.087 & 0.397 & -0.153 \\
\text{问} & 0.244 & 0.303 & 0.062 & 0.203 & 0.310 & 0.000 \\
}}_{\text{Second constituents' embeddings } }
\nonumber \\
&=&
\kbordermatrix{
      & dim1 & dim2 & dim3 & dim4 & dim5 & dim6 \\
\text{学生-生} & -0.406 & 0.188 & 0.051 & 0.221 & -0.124 & 0.388 \\
\text{学术-术} & 0.094 & -0.123 & -0.183 & -0.010 & -0.320 & -0.542 \\
\text{学者-者} & -0.078 & -0.001 & -0.151 & -0.071 & -0.226 & 0.012 \\
\text{学费-费} & -0.393 & -0.186 & -0.266 & 0.020 & -0.276 & 0.019 \\
\text{学问-问} & -0.318 & -0.280 & -0.380 & -0.114 & -0.090 & -0.030 \\
} \nonumber \\
& & \\
\bar{\bm{d}}_{\text{学}, 1}
&=&
\kbordermatrix{
      & dim1 & dim2 & dim3 & dim4 & dim5 & dim6 \\
      & -0.220 & -0.080 & -0.186 & 0.009 & -0.207 & -0.031 \\
}
\end{eqnarray*}
\caption{Illustration of the calculation of the centroid of the first constituent shift vectors of 学 (simplified data).}
\label{fig:cam_xue}
\end{sidewaysfigure}

\begin{sidewaysfigure}
\renewcommand{\kbldelim}{(}
\renewcommand{\kbrdelim}{)}
\footnotesize
\begin{eqnarray*}
{\bm{D}}_{\text{生},2}
&=&
\underbrace{
\kbordermatrix{
      & dim1 & dim2 & dim3 & dim4 & dim5 & dim6 \\
\text{医生} & \hspace*{0.8em}0.187 & -0.149 & -0.102 & -0.073 & 0.202 & -0.081 \\
\text{学生} & -0.219 & \hspace*{0.8em}0.039 & -0.051 & \hspace*{0.8em}0.148 & \hspace*{0.8em}0.078 & \hspace*{0.8em}0.307 \\
\text{人生} & -0.124 & \hspace*{0.8em}0.175 & -0.081 & -0.169 & \hspace*{0.8em}0.356 & \hspace*{0.8em}0.001 \\
\text{众生} & \hspace*{0.8em}0.158 & \hspace*{0.8em}0.038 & -0.089 & -0.108 & \hspace*{0.8em}0.211 & -0.210 \\
\text{今生} & -0.073 & -0.090 & \hspace*{0.8em}0.141 & -0.087 & \hspace*{0.8em}0.397 & -0.153 \\
\text{后生} & \hspace*{0.8em}0.244 & \hspace*{0.8em}0.303 & \hspace*{0.8em}0.062 & \hspace*{0.8em}0.203 & \hspace*{0.8em}0.310 & \hspace*{0.8em}0.000 \\
}}_{\text{Compound embeddings } }
-
\underbrace{
\kbordermatrix{
      & dim1 & dim2 & dim3 & dim4 & dim5 & dim6 \\
\text{医} & -0.030 & \hspace*{0.8em}0.052 & -0.264 & -0.179 & \hspace*{0.8em}0.036 & -0.541 \\
\text{学} & \hspace*{0.8em}0.080 & \hspace*{0.8em}0.037 & -0.240 & -0.179 & -0.015 & -0.198 \\
\text{人} & -0.466 & -0.276 & -0.125 & -0.067 & \hspace*{0.8em}0.121 & -0.134 \\
\text{众} & -0.074 & \hspace*{0.8em}0.023 & -0.318 & \hspace*{0.8em}0.089 & \hspace*{0.8em}0.220 & -0.030 \\
\text{今} & -0.219 & \hspace*{0.8em}0.039 & -0.051 & \hspace*{0.8em}0.148 & \hspace*{0.8em}0.078 & \hspace*{0.8em}0.307 \\
\text{后} & -0.124 & \hspace*{0.8em}0.175 & -0.081 & -0.169 & \hspace*{0.8em}0.356 & \hspace*{0.8em}0.001 \\
}}_{\text{First constituents' embeddings } }
\nonumber \\
&=&
\kbordermatrix{
      & dim1 & dim2 & dim3 & dim4 & dim5 & dim6 \\
\text{医生-医} & \hspace*{0.8em}0.217 & -0.201 & \hspace*{0.8em}0.162 & \hspace*{0.8em}0.106 & \hspace*{0.8em}0.166 & \hspace*{0.8em}0.460 \\
\text{学生-学} & -0.299 & \hspace*{0.8em}0.002 & \hspace*{0.8em}0.189 & \hspace*{0.8em}0.327 & \hspace*{0.8em}0.093 & \hspace*{0.8em}0.505 \\
\text{人生-人} & \hspace*{0.8em}0.342 & \hspace*{0.8em}0.451 & \hspace*{0.8em}0.044 & -0.102 & \hspace*{0.8em}0.235 & \hspace*{0.8em}0.135 \\
\text{众生-众} & \hspace*{0.8em}0.232 & \hspace*{0.8em}0.015 & \hspace*{0.8em}0.229 & -0.197 & -0.009 & -0.180 \\
\text{今生-今} & \hspace*{0.8em}0.146 & -0.129 & \hspace*{0.8em}0.192 & -0.235 & \hspace*{0.8em}0.319 & -0.460 \\
\text{后生-后} & \hspace*{0.8em}0.368 & \hspace*{0.8em}0.128 & \hspace*{0.8em}0.143 & \hspace*{0.8em}0.372 & -0.046 & -0.001 \\
}
\nonumber \\
\\
\bar{\bm{d}}_{\text{生},2}
&=&
\kbordermatrix{
      & d1 & d2 & d3 & d4 & d5 & d6 \\
      & 0.168 & 0.044 & 0.160 & 0.045 & 0.126 & 0.076 \\
}
\end{eqnarray*}
\caption{Illustration of the calculation of the centroid of the second constituent shift vectors of 生 (simplified data).}
\label{fig:cam_sheng}
\end{sidewaysfigure}

Let $\bm{d}_{\text{学,1}}^{(\text{学生})}$ denote the shift vector for the compound 学生 calculated for its first constituent:
\[
\bm{d}_{\text{学,1}}^{(\text{学生})} = \bm{s}_{\text{学生}} - \bm{s}_{\text{生}}.
\]

\noindent
The average shift vector for 学, the centroid of the $n=6$ shift vectors,  \(\bar{\bm{d}}_{\text{学},1}\), is

\begin{equation}
\bar{\bm{d}}_{\text{学},1} = \frac{1}{n} \label{eq:shift1} \sum_{i=1}^{n} \bm{d}_{\text{学,1}}^{(w_i)},
\end{equation}

\noindent
where $w_i$ is the $i$-th member of the left constituent family of 学. This centroid approximates the prototypical semantic contribution of 学 to the compounds in which it appears as the first constituent. For the second constituent of 学生, we proceed in the same way:  

\begin{equation}
\bar{\bm{d}}_{\text{生},2} = \frac{1}{n} \sum_{i=1}^{n} \bm{d}_{\text{生,2}}^{(w_i)}, \label{eq:shift2}
\end{equation}
\noindent
where $w_i$ is the $i$-th member of the right constituent family. Given these centroids of the shift vectors for the two constituent families, CAM defines the predicted meaning of the compound simply as:  

\begin{equation}
\hat{\bm{s}}_{\text{学生}} = \frac{(\bm{s}_{\text{学}} + \bar{\bm{d}}_{\text{学}, 1}) + (\bm{s}_{\text{生}} + \bar{\bm{d}}_{\text{生}, 2})}{2}.
\end{equation}

\noindent
We divide by 2 (i.e., take the average of the two constituent-based transcended embeddings) in order to constrain the length of the predicted compound embedding.  In general, given a compound $w$ with constituents $c_1$ and $c_2$, and the corresponding embeddings $\bm{s}_w, \bm{s}_{c_1}$ and $\bm{s}_{c_2}$, the predicted semantic vector for $w$ is
\begin{equation}
\hat{\bm{s}}_w = \frac{(\bm{s}_{c_1} + \bar{\bm{d}}_{c_1, 1}) + (\bm{s}_{c_2} + \bar{\bm{d}}_{c_2, 2}))}{2}.
\label{eq: compound level}
\end{equation}
The $n \times 300$ matrix $\hat{\bm{S}}$ with the predicted embeddings for $n$ compounds is defined as 
\begin{equation}
\hat{\mathbf{S}} = \frac{(\mathbf{S}_1 + \bar{\mathbf{D}}_1) + (\mathbf{S}_2 + \bar{\mathbf{D}}_2)}{2}.
\end{equation}
where $\bar{\bm{D}}_1$ and $\bar{\bm{D}}_2$ are the matrices with the corresponding shift vector centroids, as defined in (\ref{eq:shift1}) and (\ref{eq:shift2}).
Note that the row vectors of the matrices and the right hand side of the equation will have the same row vectors for compound constituents that occur in multiple compounds. 

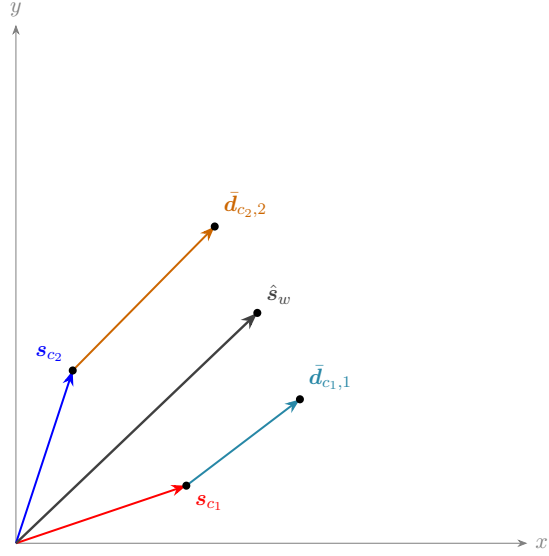
\begin{figure}[H]
\centering
\resizebox{0.45\textwidth}{0.45\textwidth}{%
\begin{tikzpicture}[line width=1pt, font=\small, >=Stealth]

\draw[->, thin, black!50] (0,0) -- (9,0) node[right] {$x$};
\draw[->, thin, black!50] (0,0) -- (0,9) node[above] {$y$};

\draw[->, red]  (0,0) -- (3,1) node[below right] {$\bm{s}_{c_1}$};
\draw[->, blue] (0,0) -- (1,3) node[above left]  {$\bm{s}_{c_2}$};

\draw[->, cyan!60!black]   (3,1) -- (5,2.5) node[above right] {$\bar{\bm{d}}_{c_1, 1}$};
\draw[->, orange!80!black] (1,3) -- (3.5,5.5) node[above right] {$\bar{\bm{d}}_{c_2,2}$};

\draw[->, very thick, gray!50!black] (0,0) -- (4.25, 4) node[above right] {$\hat{\bm{s}}_w$};

\filldraw[black] (3,1) circle (1.5pt);
\filldraw[black] (1,3) circle (1.5pt);
\filldraw[black] (5,2.5) circle (1.5pt);
\filldraw[black] (3.5,5.5) circle (1.5pt);
\filldraw[black] (4.25,4) circle (1.5pt);

\end{tikzpicture}
}
\caption{Geometric illustration of CAM. Constituent embeddings $\bm{s}_{c_1}$ and $\bm{s}_{c_2}$ act as anchor points, while centroid shift vectors $\bar{\bm{d}}_{c_1,1}$ and $\bar{\bm{d}}_{c_2,2}$ provide position-specific adjustments. The mean of the two transcended contituents  $((\bm{s}_{c_1} + \bar{\bm{d}}_{c_1,1}) + (\bm{s}_{c_2} + \bar{\bm{d}}_{c_2,2}))/2$ yields the predicted  compound representation $\hat{\bm{s}}_w$.}
\label{fig:cam_geometry_sum}
\end{figure}

Figure \ref{fig:cam_geometry_sum} provides a geometric illustration of CAM in a simplified two-dimensional semantic space. The predicted compound embedding is defined as the sum of the two constituent embeddings and the corresponding centroid shift vectors.  The constituent embeddings $\bm{s}_{c_1}$ and $\bm{s}_{c_2}$ act as anchor points, while the centroid shift vectors $\bar{\bm{d}}_{c_1, 1}$ and $\bar{\bm{d}}_{c_2,2}$ provide position-specific adjustments that are the model's best estimate of how to `transcend' the meanings of the individual constituents as independent words. 

It is worth noting that the CAM model makes use of local shifts in semantic space, and that shifts (translations) cannot be implemented with linear transformations. Therefore the CAOSS model, which uses a linear transformation, is not a general case of the CAM model.\footnote{Translations can be implemented with matrices in a higher-dimensional space using homogeneous coordinates. These matrices then represent non-linear affine transformations.}

\section{Computational experiments}

Having introduced the technical details of the CAOSS and CAM models, in this section, we report computational experiments designed to compare the predictive accuracy of the two computational accounts of semantic composition\textcolor{black}{: a learned global transformation versus local analogical generalization.}

\subsection{Materials}\label{sec:compexp}

The dataset on which we evaluated the performance of the CAOSS and CAM models was constructed from the Chinese Lexical Database (henceforth CLD)\footnote{\url{https://www.chineselexicaldatabase.com/}}, a lexical database containing lexical information for one- to four-character Chinese words written with simplified characters. This database contains 34,233 two-character words, 7,143 three-character words, and 3,355 four-character words. Among two-character words, nouns constitute the largest category (54.3\%). Nouns are even more dominant among three-character words, accounting for 89.9\% of the entries. Among four-character words, verbs constitute the largest category (44.1\%), while nouns form the second largest category (41.5\%).\footnote{ 
Part-of-speech tags in this corpus were assigned using Stanza \citep{qi2020stanza}.}

Since both CAOSS and CAM operate on constituents, and as constituent structure is not indicated in the CLD, we segmented character strings into their immediate constituents as follows.  For two-character compounds, the first and second characters were directly assigned as first and second constituents. Three-character words can be left-branching (e.g., 价目表 ({\em jia4mu4biao}, `price list', consisting of the two-character compound 价目 `price item' and the one-character constituent 表 `list'), right-branching (老朋友 ({\em lao3peng2you3}, `long-time friend',   consisting of the one-character constituent 老 (`old/long-time') and the two-character compound 朋友 (`friend'), or not branching (e.g., 对不起 ({\em dui4bu4qi3}, `sorry') and 木乃伊 ({\em mu4nai3yi1}, `mummy').  Constituent boundaries were determined based on a lookup table of two-character words. If only one two-character sequence (either the first–second or the second–third characters) existed in the two-character word dictionary, the word was segmented accordingly. When both adjacent two-character sequences of a three-character word were found in the two-character word list (15.1\% of all segmentable 3-character strings), the two-character word with the highest frequency was selected.  When neither candidate bigram existed in the two-character word dictionary, the three-character word was treated as non-segmentable. Overall, 85.7\% of the three-character words were segmentable (66.3\% left- branching, 19.4\% right-branching). The remaining 14.3\% of non-segmentable words, pose a challenge for both the CAOSS and CAM models, were excluded from further analysis.

Four-character sequences can be idioms, transliterations, or compounds.  An example of an idiomatic expression is 洛阳纸贵 ({\em luo4yang2zhi3gui4}, `widely popular'),  which is not compositionally related to the literal interpretation of paper becoming expensive in Luoyang. 奥林匹克 ({\em ao4lin2pi3ke4}, `Olympics') is an example of a transliteration. 十四行诗 ({\em shi2si4hang2shi1}, `sonnet', literally `fourteen-line poem') exemplifies the rare case of a [2+1]+1 structure. Some 4-character sequences have a balanced binary structure, without the constituents being independent words.  Examples include 一心一意 ({\em yi1xin1yi1yi4}, `wholeheartedly'), 龙凤呈祥 ({\em long2feng4\-cheng2xiang2}, `auspicious union symbolized by the dragon and phoenix'), and 含情脉脉 ({\em han2\-qing2\-mo4mo4}, `full of affection and tenderness'). After manual inspection of all 4-character sequences in the CLD, we excluded the 31 four-character words that could not be reliably segmented into two two-character constituents. The remaining four-character words (98.7\%) were segmented into two two-character constituents (e.g., 百万富翁 (\textit{bai3wan4fu4wong1}, `million-rich person', `millionaire'). 

In total, 33,844 two-character words (of which 18,422 nouns), 6,008 three-character words (of which 5,566 nouns), and 2,274 four-character words (of which 963 nouns) were included for analysis,\footnote{\label{ft:POS tag} A substantial proportion of four-character words labelled as nouns by the POS tagger do not correspond to prototypical nominal expressions. Manual inspection suggests that only 254 out of 963 four-character compounds are unambiguously tagged as nouns.  For the four-character words, the subset of 963 words labelled as nouns are largely only approximately nominal.} 
together with the corresponding Tencent embeddings \citep{song2018directional} of the words and those of their immediate constituents, using simplified Chinese characters to represent words' written forms. 

\subsection{Method}

Model performance was evaluated using nearest-neighbour retrieval, using the Pearson correlation as similarity metric. For each predicted embedding $\hat{\bm{s}}_i$, its correlation with all gold-standard TENCENT  embeddings ${\bm{s}_j}$ was computed. A prediction was considered correct if the gold-standard target embedding $\bm{s}_i$ showed the highest correlation with $\hat{\bm{s}}_i$ of all gold-standard embeddings. Accuracy was defined as the proportion of correctly retrieved target embeddings across the dataset. This metric was computed using the \texttt{eval\_SC} function in the \textbf{JudiLing} package \citep{luo2021judiling}.

Of interest to us are both how well a model predicts for known words, and how well it predicts for novel words it has not encountered during training.  The higher the accuracy of the model for known words is, the more regular and transparent the lexical system is, and the less it depends on memorization.  The more accurate model predictions are for held-out data, the more productive the compound system is.  We therefore implemented a cross-validation set-up with 30 runs in which 10\% of the words were randomly selected as held-out data and training proceeded on the remaining 90\% of the words. In this way, for both training and held-out data, a point estimate of accuracy is obtained, as well as insight into the variability around this point estimate.

In what follows, we first consider prediction accuracy evaluated with the Pearson correlation, and then also inspect how well the two models approximate the length of the compound embeddings, using the L2-norm.

\subsection{Prediction accuracy}

We first report results for the full dataset. Following this, we also report results for subsets of the data defined by length and by word category.

\subsubsection{Full data set}

\begin{figure} [htbp]
    \centering
    \includegraphics[width=0.75\linewidth]{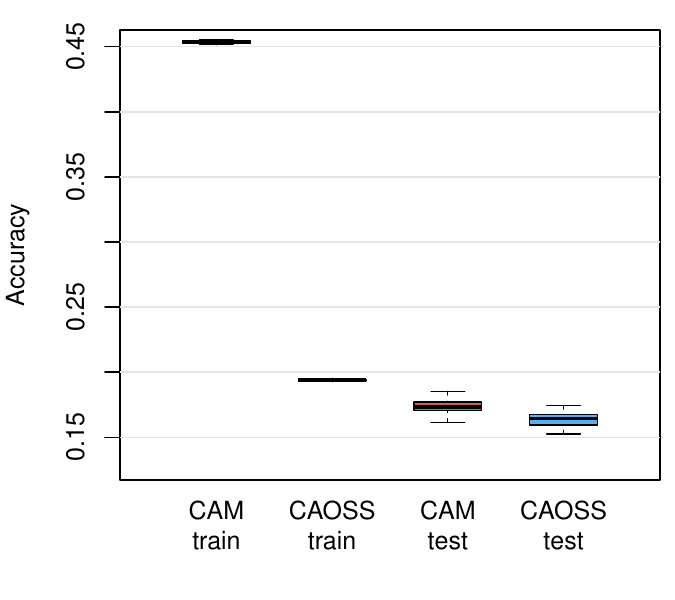}
    \caption{Prediction accuracy for CAM and CAOSS models for training and test data. For each of 30 runs, data were randomly split into training (90\%) and test (10\%) sets.}
    \label{fig:wd_all}
\end{figure}

Figure \ref{fig:wd_all} presents the results for the full dataset. On the training set, CAM achieved consistently higher prediction accuracy than CAOSS. The mean difference was 0.26, which, evaluated on all 30 pairs of training data sets, was well-supported by a t-test ($t(29) = 1167.4, p < 0.001$).  As expected, prediction accuracy decreased for both models when evaluated on the held-out datasets.  CAM outperformed CAOSS by a small but well-supported margin (means = 0.17 vs. 0.16, $t(29) = -8.99, p < 0.001$).  Comparing performance on training and test data, CAM would seem to overfit more than CAOSS, as the magnitude of the reduction in accuracy from training to testing is greater. Nevertheless, accuracy on both training and testing data is higher for CAM.

\begin{figure} [H]
    \centering
    \includegraphics[width=0.5\linewidth]{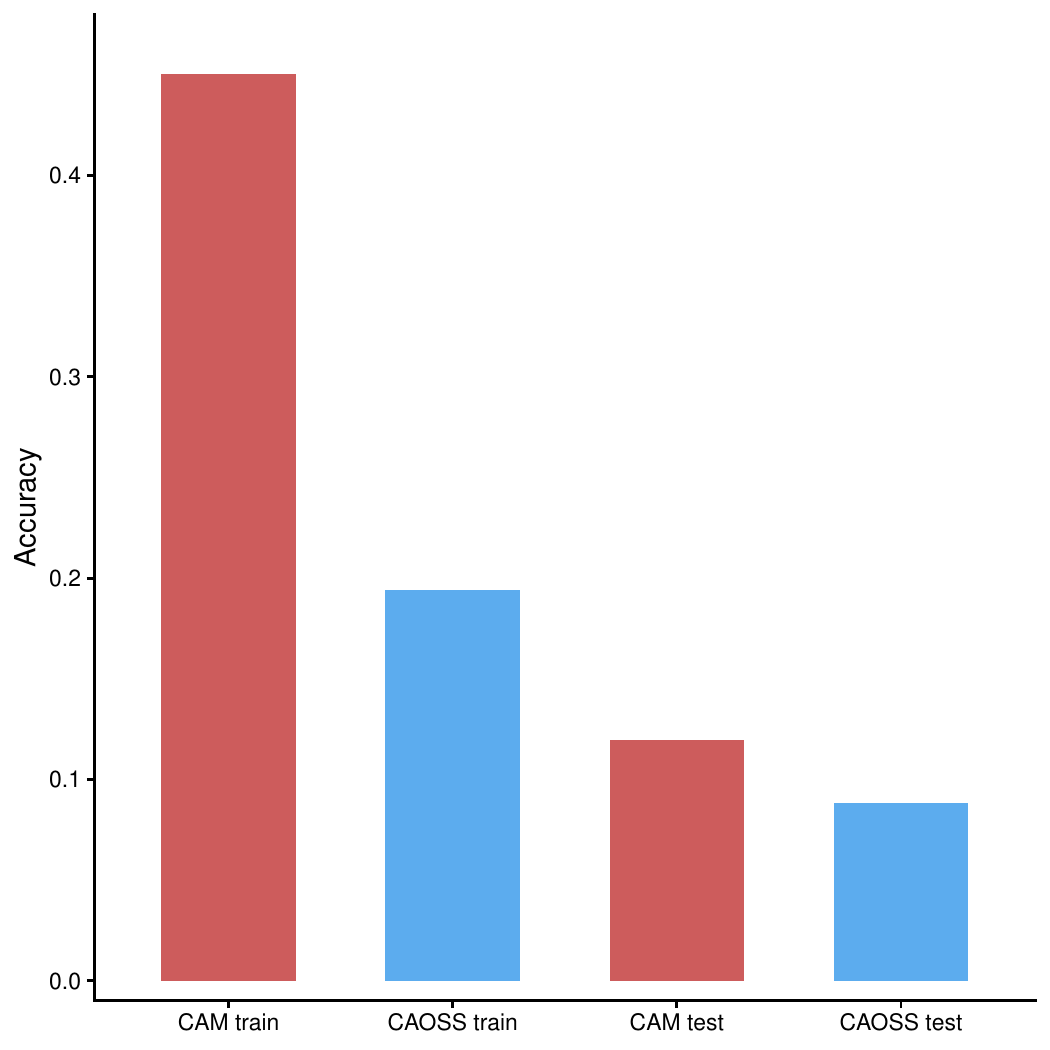}
    \caption{Prediction accuracy of CAM and CAOSS under a frequency-based split with the lowest-frequency 10\% of compounds in the test set and the remaining 90\% in the training set.}
    \label{fig:wd_all_bottom_freq}
\end{figure}

In these analyses, the selection of held-out data was completely random. As a consequence, both higher-frequency and lower-frequency words are included in the held-out datasets.  In the actual experience of language users, however, most higher-frequency words are familiar, whereas the lower-frequency compounds, which are more likely to be unfamiliar, pose challenges for interpretation.  We therefore ran a second experiment on the full dataset, in which we implemented a single split, with the training data containing the 90\% most frequent words, and the held-out data containing the 10\% lowest-frequency words. In this way, we can evaluate to what extent knowledge acquired from the more frequent compounds generalizes to the less-frequent compounds. Figure \ref{fig:wd_all_bottom_freq} shows that CAM outperformed CAOSS not only on the training set of  familiar compounds (0.45 vs. 0.19, $p < 0.0001$, proportions test), but also on the low-frequency compounds in the test set (0.12 vs. 0.09, $p < 0.0001$, proportions test). 

Considered jointly, these analyses indicate that the CAM model provides superior precision for known words, and somewhat enhanced precision for novel words, as compared to the CAOSS model: the CAM model is more productive, albeit by a small (but non-negligible) margin.

\subsubsection{Performance by word length}

\begin{figure} [H]
    \centering
    \includegraphics[width=0.9\linewidth]{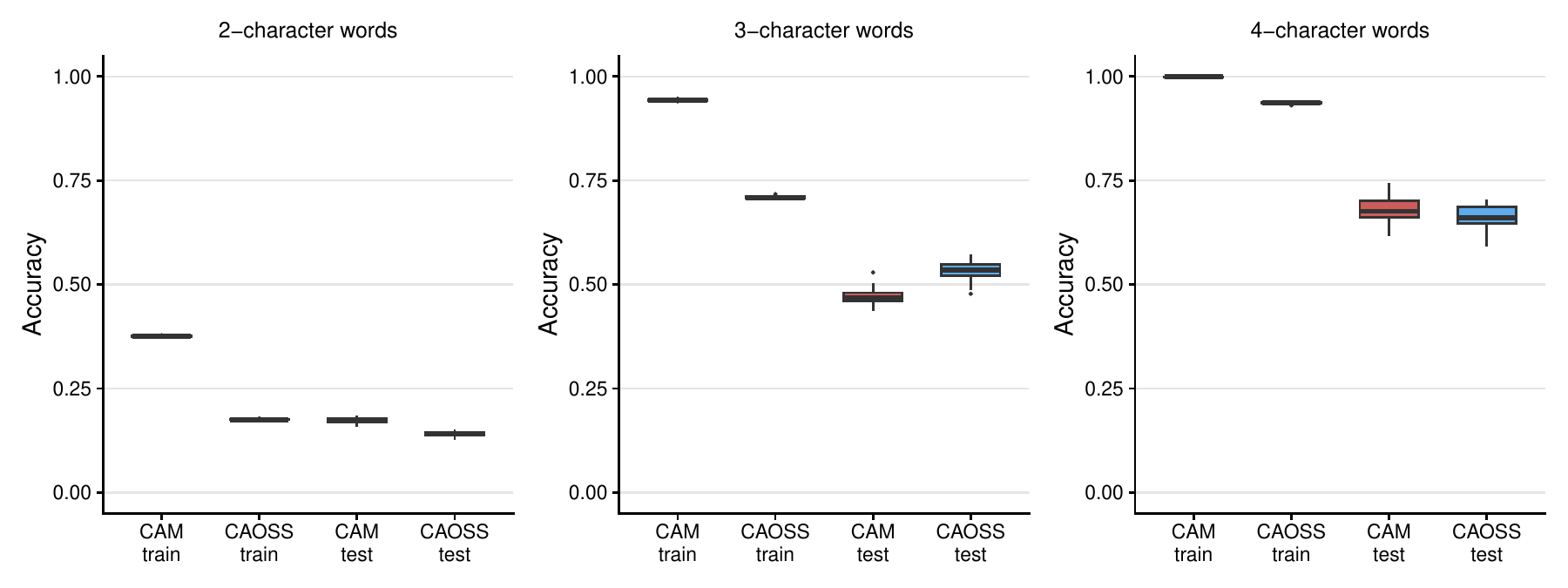}
    \caption{Boxplots of  accuracy for CAM (red) and CAOSS (blue) models trained on only the words with a given length (2, 3, and 4 characters). CAM outperformed CAOSS except for the held-out data with three-character compounds.}
    \label{fig:word length separation for all}
\end{figure}

We next split the data by word length into three subsets: a subset of two-character words (33,844), a subset with three-character words (6,008), and a subset with four-character compounds (2,274). For each of these three subsets, a CAM and a CAOSS model was constructed. Figure \ref{fig:word length separation for all} summarizes prediction accuracy of CAM and CAOSS across the three word lengths for both training and test sets. For two-character compounds, CAM (mean accuracy = 0.38 for training and 0.17 for test) outperformed CAOSS (mean accuracy = 0.18 for training and 0.14 for test) on both the training ($t(29) = 840.6, p < 0.001$) and test sets ($t(29) = 27.7, p < 0.001$). For three-character compounds, CAM achieved higher prediction accuracy than CAOSS on the training set (mean accuracy = 0.94 vs. 0.71), $t(29) = 374.36, p < 0.001$). On the test set, however, CAOSS (mean accuracy = 0.53) outperformed CAM (mean accuracy = 0.47; $t(29) = -19.76, p < 0.001$). For four-character compounds, CAM again showed higher prediction accuracy than CAOSS on both the training set (mean accuracy = 0.99 vs. 0.94; $t(29) = 136.08, p < 0.001$), and the test set (mean accuracy = 0.69 vs. 0.67; $t(29) = 3.33, p = 0.002$).

Overall, prediction accuracy increased with compound length for both CAM and CAOSS, with nearly perfect performance on the training data for CAM. For the test data, accuracy was around 17\% for two-character words, which increased to around 47\% for three-character words, and to around 68\% for four-character compounds.  Clearly, longer compounds are more semantically transparent, and offer better generalization.

However, the variability in prediction accuracy also increases with word length: In Figure~\ref{fig:word length separation for all}, the boxplots for the held-out data become wider for increasing word length.  This is likely due to increasing data sparsity.

\begin{figure} [H]
    \centering
    \includegraphics[width=0.8\linewidth]{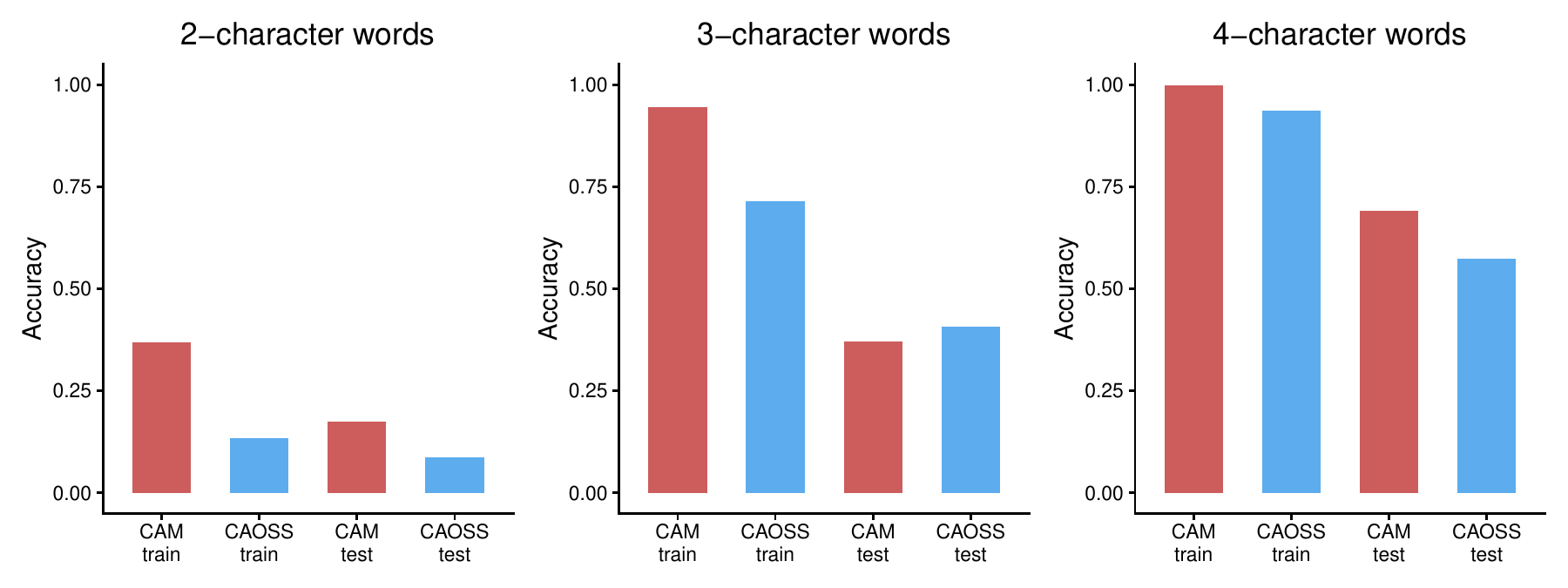}
    \caption{Prediction accuracy of CAM and CAOSS for compounds of different word lengths under a frequency-based split, with the lowest-frequency 10\% of compounds assigned to the test set and the remaining 90\% to the training set. CAM significantly outperformed CAOSS only for the held-out data of two-character compounds.}    \label{fig:bottom_freq_diff_word_length}
\end{figure}

A similar pattern of results emerged when the lowest-frequency 10\% of compounds were assigned to the test set and the remaining higher-frequency compounds were assigned to the training set (Figure \ref{fig:bottom_freq_diff_word_length}). CAM significantly outperformed CAOSS on the training sets across all three word lengths. With respect to the low-frequency  held-out data, a significant difference was present only for the two-character compounds (see Table \ref{tab: accuracy comparision arcoss word lengths}).

\begin{table}[H]
\centering
\caption{Comparison of CAM and CAOSS accuracy for training and test splits based on frequency, using proportions tests.}
\label{tab:accuracy_length_comparison}
\begin{tabular}{llcccc}
\hline
Word length & Dataset & CAM & CAOSS & $\chi^2(1)$ & $p$ \\
\hline
2-char & Train & 0.37 & 0.13 & 4455.7 & $<.0001$ \\
       & Test  & 0.17 & 0.09 & 111.26 & $<.0001$ \\
\hline
3-char & Train & 0.94 & 0.71 & 1007.8 & $<.0001$ \\
       & Test  & 0.37 & 0.41 & 1.40 & 0.24 \\
\hline
4-char & Train & 0.99 & 0.94 & 52.90 & $<.0001$ \\
       & Test  & 0.69 & 0.56 & 2.69 & 0.10 \\
\hline
\label{tab: accuracy comparision arcoss word lengths}
\end{tabular}
\end{table}

\begin{figure} [H]
    \centering
    \includegraphics[width=0.9\linewidth]{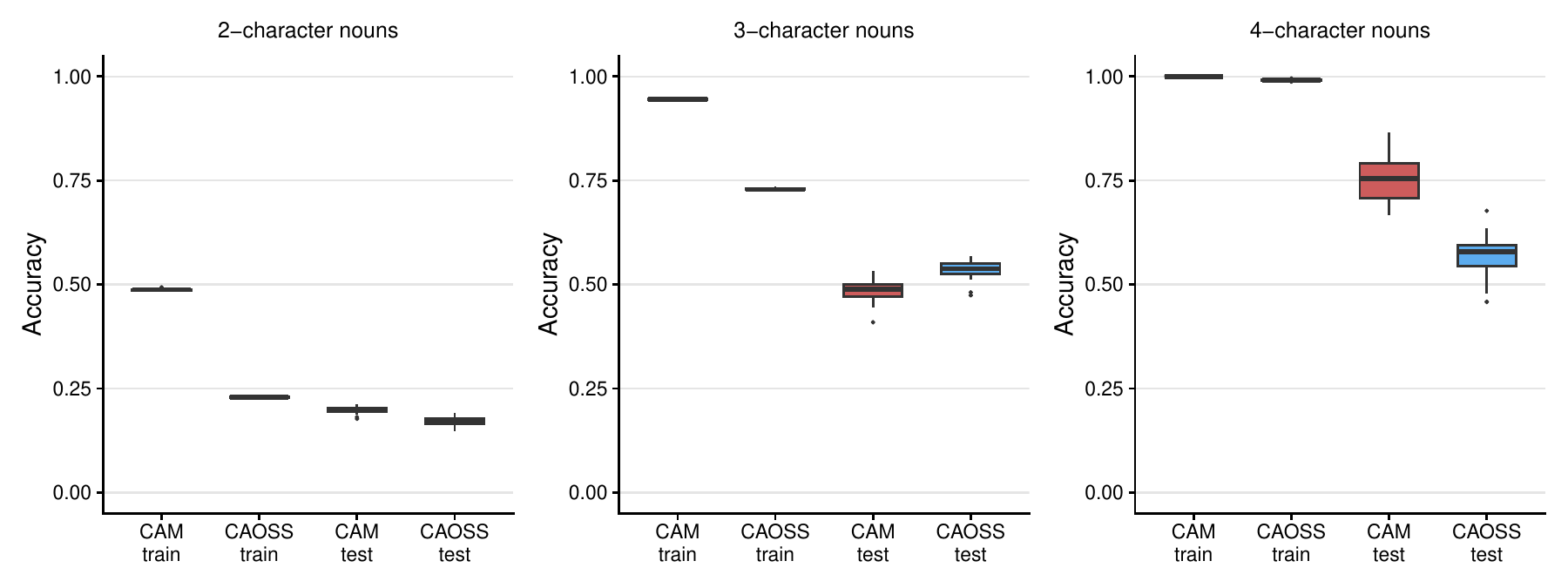}
    \caption{Boxplots for CAM (red) and CAOSS (blue) accuracies nouns by word length, for 30 cross-validation runs.}
    \label{fig:word length separation for noun}
\end{figure}

Across word lengths, the majority of compounds were nouns. Restricting the analyses to the compounds tagged as nouns again resulted in very similar results, as shown in Figure \ref{fig:word length separation for noun}, with CAOSS outperforming CAM only for the test data of three-character compounds.\footnote{Very similar results were obtained for test sets with the lower frequency words. Details are available in the supplementary materials at \url{https://osf.io/cg745/overview?view_only=07d00fe3a49440bcb130024b5443aabf}.}  Specifically, for two-character nouns, CAM (mean accuracy = 0.49 for training set and 0.20 for test) surpasses CAOSS (mean accuracy 0.23 for training and 0.17 for testing) on both training ($t(29)=658.3, p<0.001$) and test set ($t(29) = 18.1, p <0.001$).  For three-character nouns, CAM reached a higher prediction accuracy on the training set (0.95 vs. 0.73, $t(29) = 364.3, p<0.001$), but a lower accuracy on the test set (0.49 vs. 0.54, $t(29) = -13.8, p<0.001$).  For four-character nouns, CAM again outperformed CAOSS on both training set (1.0 vs. 0.99, $t(29) = 28.2, p<0.001$), and the test set (0.75 vs. 0.57), $t(29) = 18.0, p < 0.001$.  The variance in the accuracies for held-out data were quite large for the four-character compounds, which may trace back to unreliability in the tagger (see footnote~\footref{ft:POS tag}.). 
Overall, accuracies for nouns were slightly higher than the accuracies for the full datasets. 

\subsection{Vector length}

\begin{figure}[htbp]
    \centering
    \includegraphics[width=0.9\linewidth]{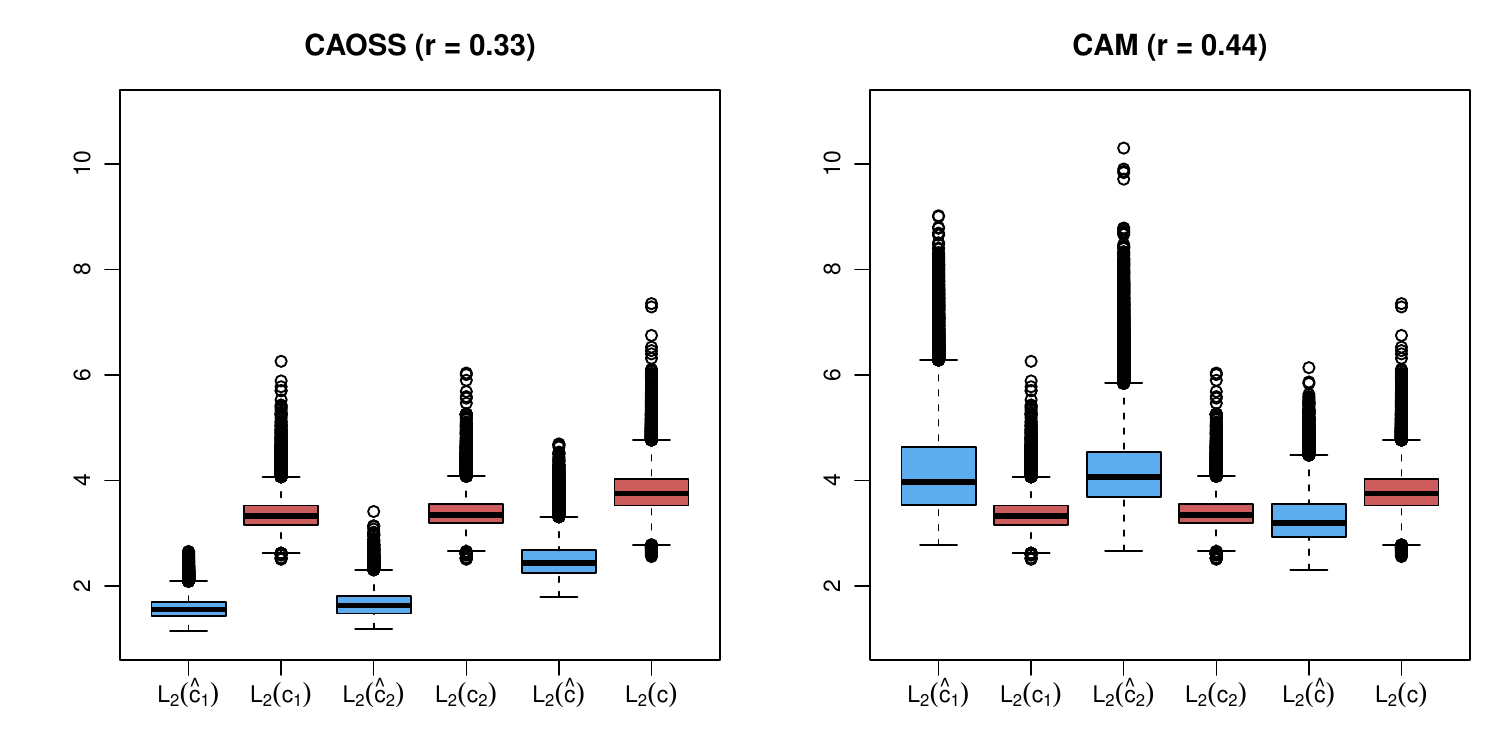}
    \caption{Distributions of the vector lengths (L2-norms) for CAOSS (left panel) and CAM (right panel), for transcended (blue) and observed (red) constituent embeddings, as well as for the predicted and observed compound vectors.}
    \label{fig:lengths_full_dataset}
\end{figure}

Figure~\ref{fig:lengths_full_dataset} visualizes the distributions of the L2-norms of the gold-standard embeddings for constituents and compounds (red boxplots) and the distributions of L2-norms of the transcended constituent vectors and the predicted compound vectors (blue boxplots) for CAOSS (left panel) and for CAM (right panel). The correlation of the predicted and observed L2-norms of the compounds was 0.33 for CAOSS and 0.44 for CAM. Apparently, the CAM model more precisely approximates compounds' L2-norms.   

It is noteworthy that for CAOSS, the L2-norms of the transcended constituent vectors as well as the L2-norms of the compounds are systematically shorter than their empirical counterparts. For the constituents, this is not problematic by itself, as the original embeddings and their transcended counterparts are not directly comparable. However, for the predicted compounds, a comparison is legitimate, and reveals that the compound embeddings predicted by CAOSS are substantially shorter than those of the corresponding gold standard vectors. 

For CAM, on the other hand, the transcended constituent vectors are somewhat longer than the original constituent embeddings, and the predicted compound vector is somewhat shorter than its gold-standard counterpart. Absolute divergences are smaller for CAM compared to CAOSS.  In sum, with respect to vector length, CAM outperforms CAOSS.

\textcolor{black}{An important property of the empirical compound embeddings is that their L2 norms are smaller than would be expected from a simple additive composition of the constituent embeddings. Under an additive model, combining two constituent vectors would generally result in a compound representation with a larger norm, {\color{black} as under independence the expectation of a sum is the sum of the expectations.} However, the empirical compound vectors are substantially shorter than this additive expectation, indicating that the semantics of compounding involve a kind of compression rather than a simple accumulation of constituent information.}

\subsection{Discussion}

A comparison of the accuracy of CAOSS and CAM on the full dataset revealed that CAM substantially outperforms CAOSS on training data, and that it also performs slightly but significantly better on held-out data. However, when the dataset is partitioned by the number of characters in a compound, results changed. First, accuracies increased substantially, approaching 1.0 for the training data with CAM and CAOSS for 4-character words. Second, for the 3-character words, CAOSS outperformed CAM on the held-out data.  The reason for these changes becomes apparent when the family sizes of the left and right constituents are considered.

For two-character compounds, median constituent family sizes are 4 for both the left and right constituents, and the percentage of constituents that occur only in 1 compound is 28\% for both the left and right character.  However, for 3-constituent words, the constituent families are extremely small: for the left constituent, the mean and median are 1.6 and 1 respectively, and for the right constituent, mean and median are 6.4 and 2.  The percentages of constituents that only occur in one compound (henceforth `hermits') are 70.1\% for the first constituent and 49.1\% for the second constituent.  These numbers are even more extreme for the 4-character compounds: 93\% of the left constituents and 89\% of right constituents occur in a single compound only. 

For compounds with hermits, analogical generalization becomes impossible. Consider again equation (\ref{eq: compound level}),  repeated here for convenience:
\begin{equation*}
\hat{\bm{s}}_w = ((\bm{s}_{c_1} + \bar{\bm{d}}_{c_1, 1}) + (\bm{s}_{c_2} + \bar{\bm{d}}_{c_2, 2}))/2.
\end{equation*}
When both $c_1$ and $c_2$ are hermits, this equation reduces to 
\begin{equation*}
\hat{\bm{s}}_w = ((\bm{s}_{c_1} + \bm{d}_{c_1, 1}) + (\bm{s}_{c_2} +\bm{d}_{c_2, 2}))/2
\end{equation*}
where
\begin{equation*}
\bm{d}_{c_1,1}=\bm{s}_w-\bm{s}_{c_2},\qquad
\bm{d}_{c_2,2}=\bm{s}_w-\bm{s}_{c_1}.
\end{equation*}
Substituting these expressions yields:
\begin{equation*}
\hat{\bm{s}}_w =
\frac{(\bm{s}_{c_1}+\bm{s}_w-\bm{s}_{c_2})
+(\bm{s}_{c_2}+\bm{s}_w-\bm{s}_{c_1})}{2}
=\bm{s}_w.
\end{equation*}
for training data, and to  
\begin{equation*}
\hat{\bm{s}}_w = (\bm{s}_{c_1} + \bm{s}_{c_2})/2 
\end{equation*}
for test data: for the held-out data, the embedding of the compound is not available, and hence no shift vectors can be calculated.  In other words, for held-out three and four-character compounds, no true analogy is possible and the CAM model reduces to summation of the constituent vectors. 

\begin{figure}
    \centering
    \includegraphics[width=0.32\linewidth]{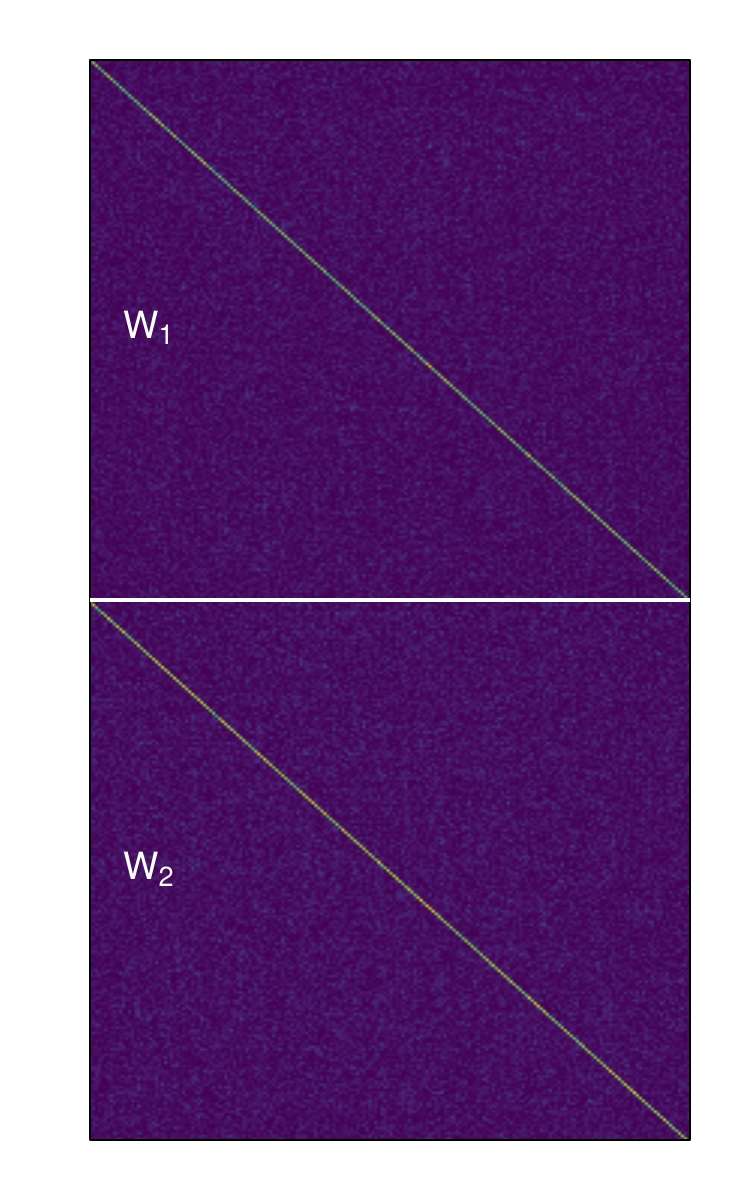}
    \caption{
    Absolute-value heatmap of the global CAOSS transformation matrix $\bm{W}$ for two-character Mandarin compounds. The two blocks correspond to the constituent-specific sub-matrices $\bm{W}_1$ and $\bm{W}_2$. Both matrices are diagonally dominant, with brighter diagonal entries indicating larger transformation weights. The diagonal elements account for approximately 72.8\% of the Frobenius norm of $\bm{W}$.}
    \label{fig:mandarin2}
\end{figure}

The CAOSS model is also challenged by the data sparsity for longer compounds. Figure~\ref{fig:mandarin2} illustrates that for two-character compounds, the submatrices $\bm{W}_1$ and $\bm{W}_2$ are rather similar to diagonal matrices. The proportion of the Frobenius norm of $\bm{W}$ that is contributed by the diagonal elements of $\bm{W}_1$ and $\bm{W}_2$ is roughly 72.8\%.\footnote{For a dataset with 4817 English compounds, this proportion is much smaller: 9.6\%.}   For the three and four-character compounds, the proportional contribution of the diagonal elements decreases slightly to 65.4\% and 59.1\%.  Clearly, for longer compounds, off-diagonal elements are more important, and make more precise prediction possible for compounds that likely are more transparent.   

In the light of these considerations, it is not straightforward to evaluate the differences in the performance of the CAM and CAOSS models on held-out data.  The better performance of CAOSS for 3-character words and the better performance of CAM for 4-character words may be just a coincidence that is contingent on the longer words that happened to be included in the CLD. It is also possible that an asymmetry in the constituent family sizes is at issue. Denoting the left and right constituent family sizes, excluding the target compounds themselves, by $V_1$ and $V_2$ respectively, an index of asymmetry for all compounds of a given length is
\begin{equation}
    \text{asym} = \text{median}(\log(V_1+1) - \log(V_2+1)).
\end{equation}
For the 2-character words, this index is 0.97, for the 3-character words it is 1.87, and for the four-character words it is basically 0. As this asymmetry index is the greatest for the 3-character compounds, it is conceivable that the analogy model works less well for held-out data when constituent family sizes are unbalanced.

Finally, consider the performance of CAM and CAOSS for training data. Whereas from a machine learning perspective, generalization to unseen data is a critical desideratum, from a cognitive perspective, accuracy on training data is equally informative. If performance on training data is very low, \textcolor{black}{this suggests that the semantic regularities exploitable by the present model are limited for the compounds that speakers frequently produce and listeners routinely understand. In this case, lexical processing is likely to depend more heavily on rote learning and memorization. Conversely, the better a model captures the representations of common words, the more it suggests that regularities in the semantic space can support lexical processing, thereby reducing the burden on memory. From this perspective, the CAM model provides a better account of the structure present in the training data than the CAOSS model.}

In the next section, we address how well the CAM and CAOSS models perform as models of lexical processing in the visual lexical decision task. 

\section{Predicting visual lexical decision response latencies}

\subsection{Data}

Data were extracted from the MELD-SCH lexical decision dataset \citep{tsang2018meld}, which comprises response times for 1,020 one-character, 10,022 two-character, 949 three-character, and 587 four-character words written with simplified Chinese characters. Lexical decision latencies were collected from 504 native Chinese speakers. We restricted our analysis to the 10,022 two-character compounds. We discarded items with error rates greater than 30\%, which left us with 9,627 words.  Intersecting this dataset with the words in our computational modeling dataset resulted in a final dataset with 8,752 two-character compounds.  

The predictors that we considered in the regression analyses included both standard predictors and model-based predictors. Word frequency counts were obtained from the SUBTLEX-CH database \citep{cai2010subtlex}\textsuperscript{8}, and log-transformed ($f(w)$).  Left and right constituent family size were calculated from the same resource, and also log-transformed ($V(c_1), V(c_2)$).  For discussion of these measures (and many others) in the context of the lexical decision task, see, e.g., 
\citep{cui2021compound,kuperman2009reading, hsieh2024beyond,li2026accessing,Whaley:78,li2015role,Baayen:Wurm:Aycock:2007,Baayen:ML:2011}.

From both the CAOSS model and CAM model, we derived three model-based predictors. 
One measure probes how successful a model is in predicting the meaning of the compound, the gold-standard ground-truth embedding. For the other two measures, we adopt the terminology of \citet{hsieh2024beyond}, and refer to these as `proximity' measures.  They assess to what extent the transcended constituent meanings are similar to the predicted meaning of the compound. The transcended constituent meanings for the CAOSS model have already been defined (see equations \ref{eq:caoss_transc1} and \ref{eq:caoss_transc2}). For the CAM model,  the transcended first constituent embedding ($\hat{\bm{s}}_{c_1}$) is $\bm{s}_{c_1} + \bar{\bm{d}}_{c_1, 1}$ and the transcended second constituent embedding ($\hat{\bm{s}}_{c_2}$) is $\bm{s}_{c_2} + \bar{\bm{d}}_{c_2, 2}$. In other words, we rewrite equation (\ref{eq: compound level}) as follows:
\begin{eqnarray*}
\hat{\bm{s}}_w &=& ((\bm{s}_{c_1} + \bar{\bm{d}}_{c_1, 1}) + (\bm{s}_{c_2} + \bar{\bm{d}}_{c_2, 2}))/2 \\
 &=& (\hat{\bm{s}}_{c_1} + \hat{\bm{s}}_{c_2})/2.
\end{eqnarray*}
For the model-based measures, we make use of the following notation, where cos means cosine similarity of embeddings:

\begin{center}

\begin{minipage}[t]{.3\textwidth}
\begin{align*}
\text{similarity of}\ \hat{\mathbf{s}}_{c_1}  \text{and}\ \hat{\mathbf{s}} & \\
\text{similarity of}\ \hat{\mathbf{s}}_{c_2}  \text{and}\ \hat{\mathbf{s}} & \\
\hat{\mathbf{s}}\ \text{and ground truth}\;  \mathbf{s} & \quad 
\end{align*}%
\end{minipage}%
\begin{minipage}[t]{.35\textwidth}
\centering
CAM
\begin{align*}
P^{CAM}_{c1}
&=
\cos\left(
\hat{\mathbf{s}}_{c1}^{CAM},
\hat{\mathbf{s}}_{w}^{CAM}
\right)
\\
P^{CAM}_{c2}
&=
\cos\left(
\hat{\mathbf{s}}_{c2}^{CAM},
\hat{\mathbf{s}}_{w}^{CAM}
\right)
\\
S^{CAM}_{C}
&=
\cos\left(
\hat{\mathbf{s}}_{w}^{CAM},
\mathbf{s}_{w}
\right)
\end{align*}

\end{minipage}
\begin{minipage}[t]{.35\textwidth}
\centering
CAOSS
\begin{align*}
P^{CAOSS}_{c1}
&=
\cos\left(
\hat{\mathbf{s}}_{c1}^{CAOSS},
\hat{\mathbf{s}}_{w}^{CAOSS}
\right)
\\
P^{CAOSS}_{c2}
&=
\cos\left(
\hat{\mathbf{s}}_{c2}^{CAOSS},
\hat{\mathbf{s}}_{w}^{CAOSS}
\right)
\\
S^{CAOSS}_{C}
&=
\cos\left(
\hat{\mathbf{s}}_{w}^{CAOSS},
\mathbf{s}_{w}
\right)
\end{align*}
\end{minipage}
\captionof{table}{Notation for the two proximity measures and the cosine similarity measure for CAM (left) and CAOSS (right).}
\label{tab:proximity_notation}
\end{center}

\noindent
We made use of the generalized additive model \citep[GAM][]{Wood:2017} to analyse the response latencies, using the implementation in the \textbf{mgcv} package for \texttt{R}.  Response times were inversely transformed (-1000/RT) before being entered into the model, based on a Box–Cox analysis and inspection of the quantile-quantile plot.

\subsection{Results}

We first report the results for the CAOSS model, and then turn to consider the performance of the CAM model.

\subsubsection{CAOSS model}

\begin{figure}[htbp]
    \centering
\includegraphics[width=1.0\linewidth]{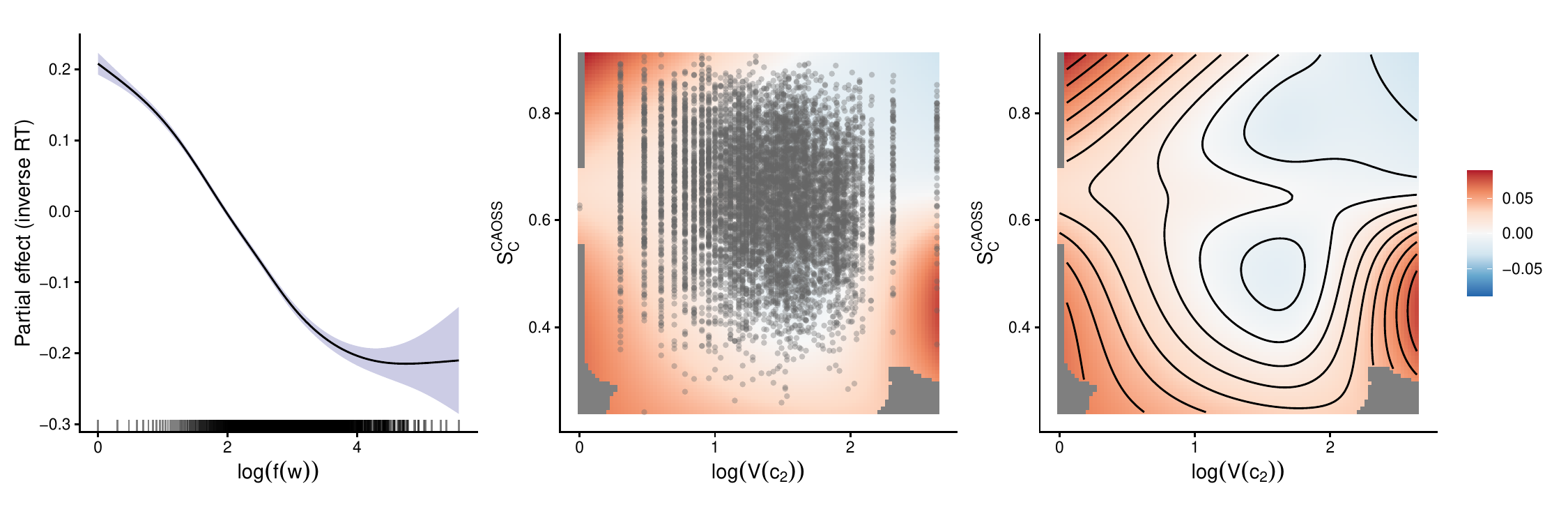}
    \caption{
Partial effects of the smooth terms in the GAM fitted to visual lexical decision latencies using predictors derived from the CAOSS model. The left panel shows the nonlinear effect of log-transformed word frequency, with the effect levelling off at higher frequency levels. The middle panel displays the observed distribution of the two predictors. The right panel shows the contour plot of the tensor smooth for the interaction between right constituent family size and cosine similarity. Colours indicate predicted response latencies, with darker blue regions representing shorter latencies and red regions representing longer latencies.
}

    \label{fig:caoss_gam2}
\end{figure}

The following GAM with CAOSS-based predictors provided an optimal fit to the data: 
\begin{equation}
\text{RT} \sim \text{s}(f(w)) + {P}_{c1}^{CAOSS}+{P}_{c2}^{CAOSS} + \text{te}(V(c_2), S_C^{CAOSS})),\label{eq:gam_caoss}
\end{equation}
(AIC=\mbox{-11570.10}).
This model provides a substantially better fit to the data than the linear model that \citet{hsieh2025compositional} proposed for their dataset:
\begin{equation}
    \text{RT} \sim f(w) + V(c_1) + V(c_2) +  P_{c1}^{CAOSS} + P_{c2}^{CAOSS},\label{eq:gam_caoss_xie}
\end{equation}
(AIC = -11456.69). Table~\ref{tab:gam_caoss} provides a summary of the GAM (\ref{eq:gam_caoss_xie}).
The two proximity measures have linear effects with negative slopes.  The magnitude of the slope of $P_{c2}^{CAOSS}$ is more than twice that of $P_{c1}^{CAOSS}$. 
Figure~\ref{fig:caoss_gam2} visualizes the partial effects of the nonlinear terms in the model. Greater word frequencies predict shorter response times; however, this effect levels off for higher-frequency words (left panel). The middle panel shows the observed distribution of the two predictors involved in the tensor product smooth, namely right constituent family size ($V(c_2)$) and the cosine of the angle between the predicted and observed compound embeddings ($S_C^{CAOSS}$). The right panel presents the corresponding contour plot of the tensor smooth. The interaction indicates that the effect of $S_C^{CAOSS}$ depends on right constituent family size. When right constituent family size is low, changes in cosine similarity are associated with relatively small and non-monotonic changes in predicted response latencies. In contrast, when right constituent family size is high, increasing cosine similarity is associated with decreasing response latencies. The shortest predicted response latencies occur for medium-to-high levels of right constituent family size and across much of the range of $S_C^{CAOSS}$. Table~\ref{tab:deltaAIC_caoss} lists the increase in AIC when a predictor (or pair of predictors, in the case of the proximity measures) are withheld from the model specification. Unsurprisingly, word frequency has the greatest variable importance, followed by the family size of the second constituent. Second constituent proximity is associated with a medium variable importance. By contrast, the variable importance of the proximity measure for the first constituent is low, and removing both proximities from the model leads to a change in AIC that is very similar to that of removing just $P_{c2}^{CAOSS}$. The similarity of the predicted compound embedding and its gold-standard counterpart also has a small variable importance.

\begin{table}[ht]
\centering
\caption{Summary of a GAM fitted to visual lexical decision latencies with predictors derived from the CAOSS model  (AIC: -11570.1).} 
\begin{tabular}{lrrrr}
   \hline
A. parametric coefficients & Estimate & Std. Error & t-value & p-value \\ 
  Intercept & -1.1593 & 0.0333 & -34.7889 & $<$ 0.0001 \\ 
  ${P}_{c1}^{CAOSS}$ & -0.0727 & 0.0226 & -3.2211 & 0.0013 \\ 
  ${P}_{c2}^{CAOSS}$ & -0.1870 & 0.0312 & -5.9939 & $<$ 0.0001 \\ 
   \hline
B. smooth terms & edf & Ref.df   & F-value & p-value \\ 
  $\mathrm{s}(\log f(w))$ & 5.3077 & 6.3023 & 933.7357 & $<$ 0.0001 \\
  \text{te}($V(c_2), S_C^{CAOSS}$) & 13.1744 & 15.6219 & 5.1673 & $<$ 0.0001 \\ \hline 
\end{tabular}

\label{tab:gam_caoss}

\end{table}

\begin{table}[htbp]
\centering
\begin{tabular}{lrrr} \hline
                   &       df   &     AIC    &   $\Delta$ AIC \\ \hline
$f(w)$             & 30.4 & -7075.4 &4494.7 \\
$V(c_2)$           & 19.4 &-11512.7 &  57.4 \\
$P_{c2}^{CAOSS}$   & 23.3 &-11537.2 &  32.9 \\
$P_{c1}^{CAOSS},P_{c2}^{CAOSS}$ & 22.8 &-11537.3 &  32.8 \\
$S_C^{CAOSS}$      & 18.5 &-11560.3 &   9.8 \\
$P_{c1}^{CAOSS}$   & 27.6 &-11561.2 &   8.9 \\ \hline
\end{tabular}
\caption{Variable importance, estimated by the change in AIC when a predictor is removed from the GAM model that is informed by CAOSS-based predictors. (\ref{eq:gam_caoss}).}
\label{tab:deltaAIC_caoss}
\end{table}

\begin{figure}[htbp]
\centering
\begin{minipage}{0.5\textwidth}
\centering
    \resizebox{1.5in}{1.5in}{%
\begin{tikzpicture}
\draw[blue,-latex,line width=1pt] (1,1)--(2,4) node[midway,above,sloped] () {$||\hat{\bm{s}}_1||$};
\draw[red,-latex,line width=1pt] (2,4)--(5,5) node[midway,above,sloped] () {$||\hat{\bm{s}}_2||$};
\draw[black,-latex,line width=1pt] (1,1)--(5,5) node[midway,below,sloped] () 
   {$ ||\hat{\bm{s}}_1|| \cos \alpha + ||\hat{\bm{s}}_2|| \cos \beta$};

\draw[color=gray,-] (2,4) to (3,3);
\node [blue] at (1.3,1.5) {$\alpha$};
\node [red] at (4.2,4.45) {$\beta$};
\end{tikzpicture}
}
\end{minipage}%
\begin{minipage}{0.5\textwidth}
\includegraphics[width=1.0\textwidth]{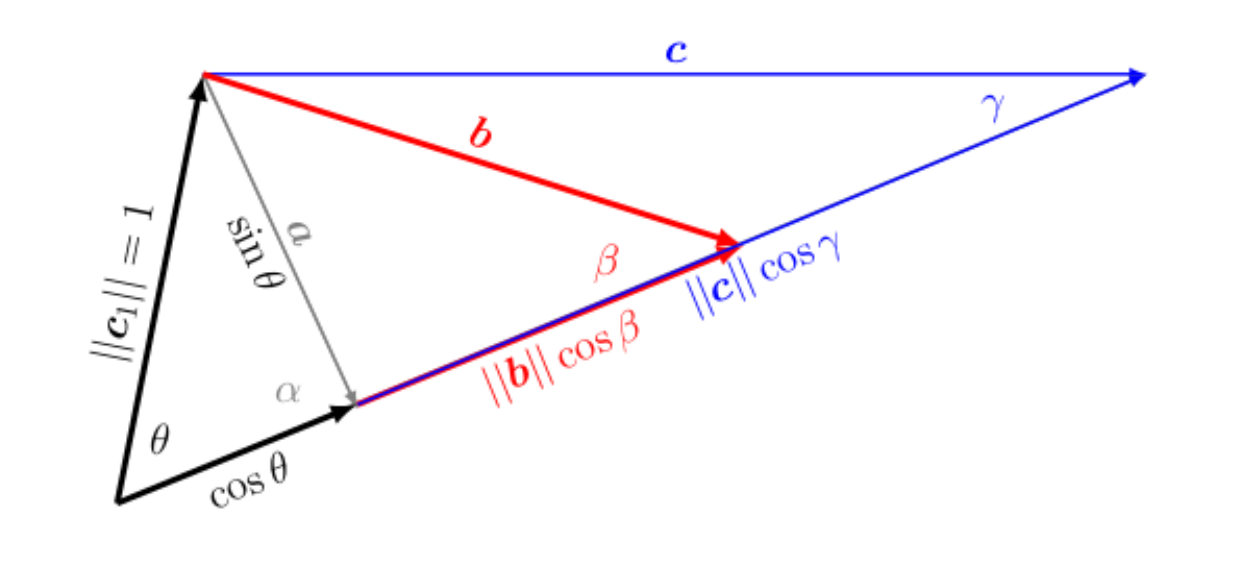}
\end{minipage}%
    
\caption{The geometry of CAOSS proximity measures. Left panel: transcended vectors of equal length; right panel: transcended vectors of different lengths.}
\label{fig:drawings}    
\end{figure}
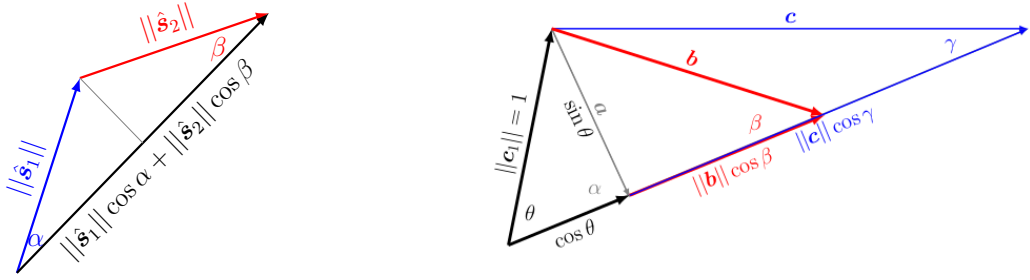

Although the proximity measures appear to assess semantic similarity, upon closer inspection it turns out that what they actually assess is their contribution to the L2-norm (length) of the compound vector.  To see this, first consider the left panel of  Figure~\ref{fig:drawings}, which presents two transcended constituent vectors, both with the same length, and the vector of their sum. The length of the compound vector is $||\hat{\mathbf{s}}_1|| \cos \alpha + ||\hat{\mathbf{s}}_2|| \cos \beta$. Since $\alpha=\beta$, in this example, the length of the compound vector is simply $2||\hat{\mathbf{s}}_1|| \cos \alpha$.  The right panel of Figure~\ref{fig:drawings} illustrates the situation in which the two transcended embeddings have different lengths. For ease of exposition, we define the first transcended vector (denoted by $\bm{c}_1$) to have unit length. Its contribution to the length of the compound vector is equal to $\cos \theta$.  The shortest possible second transcended vector is $\bm{a}$, which has length $\sin \theta$. As it is orthogonal to the compound vector, its contribution to the length of the compound vector is 0.  This might be a reasonable characterization for compounds such as \textit{cranberry}, in which the first constituent is semantically inert. For the longer vectors $\bm{b}$ and $\bm{c}$, their contributions to the length of the compound vector is $||\bm{b}||\cos \beta$ and $||\bm{c}||\cos \gamma$ respectively. The smaller the angle of the transcended second constituent vector with the predicted compound embedding is, the longer that constituent's vector must be. In other words, an increase in cosine similarity (proximity) goes hand in hand with an increase in the contribution to the length of the predicted compound embedding.

Thus, the length of the predicted  compound embedding is a weighted sum of the two cosines, in the general case $||\hat{\mathbf{s}}_1|| \cos \alpha + ||\hat{\mathbf{s}}_2|| \cos \beta$. Interestingly, in the GAM for CAOSS, the two cosines emerge with linear effects. Their contribution to the predicted response times is also a weighted sum.  If the two regression weights are calibrating for the lengths of the transcended embeddings, then the visual lexical decision latencies are in fact co-determined by the length of the predicted compound vector.  This possibility is supported by the fact that the length of the second transcended vector is greater than that of the first transcended vector (1.55 vs. 1.70, $t(16889)=-50.3, p < 0.0001$), which fits well with the greater magnitude of the regression weight for the latter in the GAM.

Following up on this possibility, we replaced the two proximity measures by the L2-norm $\hat{L}2(w)$ of the predicted compound vector.\footnote{For the L1-norm and L2-norm of embeddings as predictors of visual lexical decision times, see  \citet{Milin:Feldman:Ramscar:Hendrix:Baayen:2017}.}  The AIC of this model improved upon model (\ref{eq:gam_caoss}), reducing the AIC from -11570.10 to -11573.05. When the predicted L2-norm is replaced by the L2-norm of the gold-standard compound, the AIC reduces substantially to -11631.9 (61.8 AIC units).  These observations lead to the conclusion that the proximities of the CAOSS model considered jointly provide  estimates of a compound's L2-norm.

\subsubsection{CAM}

\begin{figure}[htbp]
    \centering
    \includegraphics[width=1.0\linewidth]{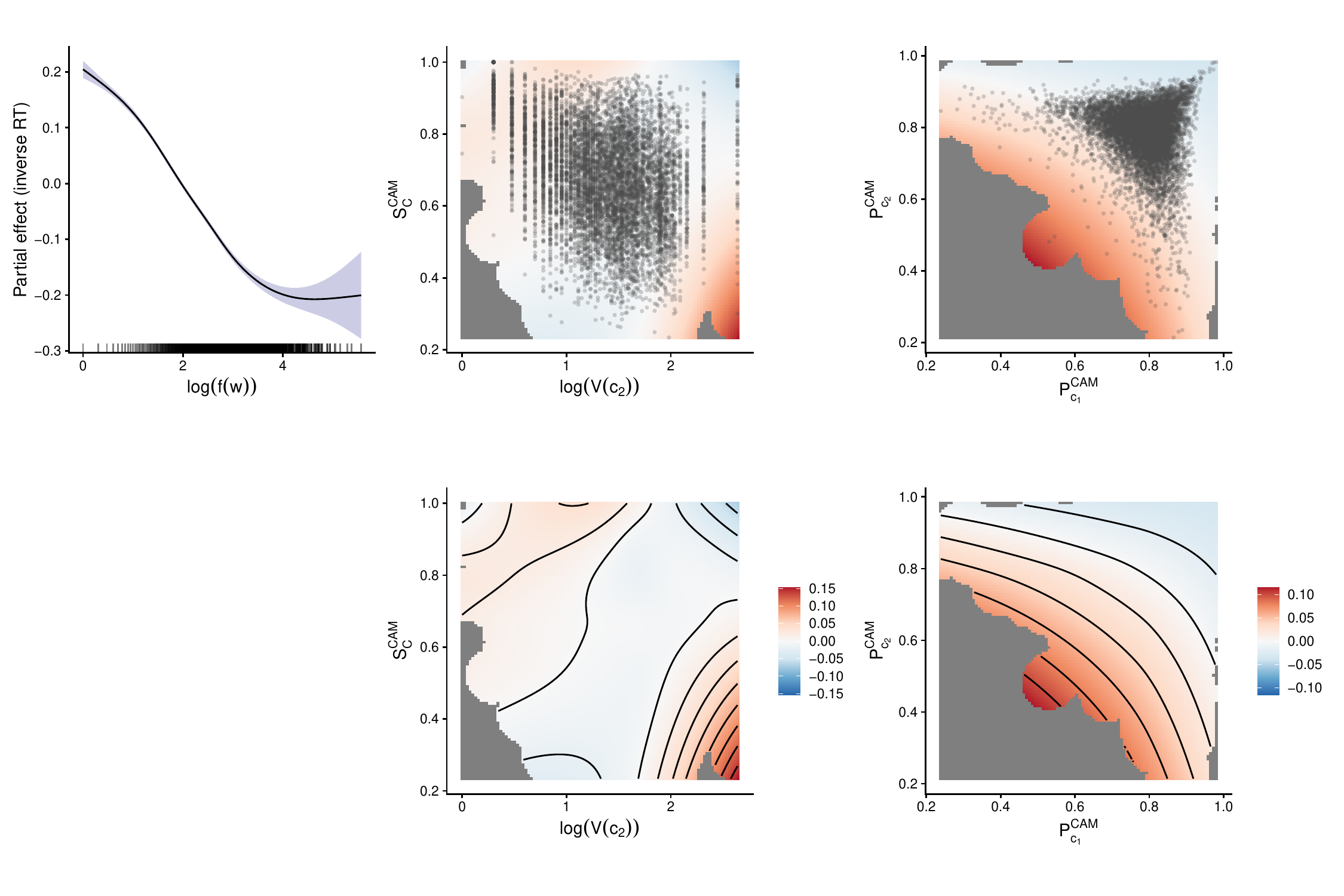}
    \caption{Partial effects of the smooth terms in a GAM fitted to visual lexical decision latencies using predictors derived from the CAM model. The first panel in first row shows the nonlinear effect of log-transformed word frequency. The middle and right panels in the first row display the observed distributions of the predictors involved in the two tensor-product smooths: right constituent family size ($V(c_2)$) with cosine similarity ($S_C^{CAM}$), and the proximity measures of the first and second constituents ($P_{c_1}^{CAM}$ and $P_{c_2}^{CAM}$), respectively. The second row presents the corresponding contour plots of the tensor smooths. Colours indicate predicted response latencies, with bluer shades representing shorter response latencies and lighter/redder shades representing longer response latencies.}
    \label{fig:cam_gam2}
\end{figure}

The optimal GAM when using predictors grounded in CAM, 
\begin{equation}
\text{RT} \sim \text{s}(f(w)) + \text{te}({P}_{c1}^{CAM}+{P}_{c2}^{CAM}) + \text{te}(V(c_2), S_C^{CAM})) \label{eq:gam_cam} 
\end{equation}
is similar to that for the CAOSS model, but the two proximity measures interact, requiring a tensor product smooth.  Table~\ref{tab:gam_cam} presents the model summary, and Figure~\ref{fig:cam_gam2} the corresponding partial effects. The partial effect of word frequency is very similar to that of the GAM for CAOSS.  The interaction of second constituent family size ($V(c_2)$) and the cosine similarity of the predicted and observed compound embeddings ($S_C^{CAM}$) suggests a cross-over interaction (center panels). For low values of $V(c_2)$, response latencies increase with $S_C^{CAM}$. For higher second constituent family sizes, this effect reverses.  The interaction of the two proximities is more regular, and indicates that response latencies decrease most when both proximities have large values.

\begin{table}[ht]
\centering
\caption{Summary of a GAM fitted to visual lexical decision latencies with predictors grounded in CAM (AIC: -11597.45).} 
\begin{tabular}{lrrrr}
   \hline
A. parametric coefficients & Estimate & Std. Error & t-value & p-value \\ 
  Intercept & -1.3555 & 0.0013 & -1018.1199 & $<$ 0.0001 \\ 
   \hline
B. smooth terms & edf & Ref.df & F-value & p-value \\ 
  $\text{s}(\log(f(w))$ & 5.4745 & 6.4724 & 852.9476 & $<$ 0.0001 \\ 
  $\text{te}(V(c_2), S_C^{CAM}$ & 14.5436 & 17.2955 & 3.4391 & $<$ 0.0001 \\ 
  $\text{te}(P_{c1}^{CAM}, Pr_{c2}^{CAM})$ & 3.8423 & 4.5405 & 17.2632 & $<$ 0.0001 \\ 
   \hline
\end{tabular}

\label{tab:gam_cam}
\end{table}

\begin{table}[ht]
\centering
\caption{Variable importance, estimated by the change in AIC when a predictor is removed from the GAM model based on CAM predictors (\ref{eq:gam_cam}).}
\begin{tabular}{lrrr} \hline
                                             & df   & AIC      & $\Delta$ AIC  \\ \hline
  $f(w)$                                     & 25.6 &  -7333.5 & 4264.0 \\ 
  $P_{c1}^{CAM},P_{c2}^{CAM}$                & 22.7 & -11527.9 & 69.6   \\ 
  $P_{c1}^{CAM}$                             & 23.6 & -11538.9 & 58.6   \\ 
  $V(c_2)$                                   & 14.1 & -11567.1 & 30.4   \\ 
  $S_C^{CAM}$                                & 19.4 & -11584.9 & 12.5   \\ 
  $P_{c2}^{CAM}$                             & 29.1 & -11589.8 & 7.7    \\ \hline
\end{tabular}

\label{tab:deltaAIC_cam}
\end{table}

Table~\ref{tab:deltaAIC_cam} reports the variable importances for the different predictors. Again, unsurprisingly, word frequency has the greatest variable importance. Removing both proximity measures leads to a strong increase in AIC. For CAM, the proximity of the first constituent is much more important than that of the second proximity, reversing the pattern observed for CAOSS.  The variable importance of $S_C^{CAM}$ is modest.  

Given that the proximity measures based on CAOSS approximate the L2-norm of the (predicted) compound, we also checked whether replacing the CAM proximities by the L2-norm of the predicted compound vector would lead to an improved fit. This was not the case: the AIC increased from  -11597.45 to -11552.12.  However, when the L2-norm of the gold standard compound embedding was used, the AIC decreased from -11597.45 to -11620.93 (23.5 AIC units).  This indicates that the proximities of the CAM model are not straightforward estimators of compound length, a conclusion that dovetails well with the finding that in the statistical model, the CAM proximities are not additive but interact. 

\subsubsection{CAOSS and CAM compared}

The goal of the CAOSS and CAM models is to provide predictions for compounds' meanings.  When predictors derived from these models are pitted against visual lexical decision latencies, the optimal model with CAOSS predictors, 
\begin{equation*}
\text{RT} \sim \text{s}(f(w)) + ||\hat{\bm{s}}_w^{CAOSS}|| + \text{te}(V(c_2), S_C^{CAOSS})), 
\end{equation*}
comes with AIC = -11573.05, whereas the optimal CAM model,
\begin{equation*}
\text{RT} \sim \text{s}(f(w)) + \text{te}({P}_{c1}^{CAM}+{P}_{c2}^{CAM}) + \text{te}(V(c_2), S_C^{CAM})), 
\end{equation*}
comes with with AIC = -11597.45. This suggests that the CAM model has an advantage over CAOSS for predicting visual lexical decision latencies.  However, as results will vary depending on the choice of word frequencies, the kind of embeddings, and the inclusion of covariates, we limit our conclusion to the observation that the CAM model is a competitive alternative to the CAOSS model.

\begin{figure}[htbp]
    \centering
    \includegraphics[width=0.7\linewidth]{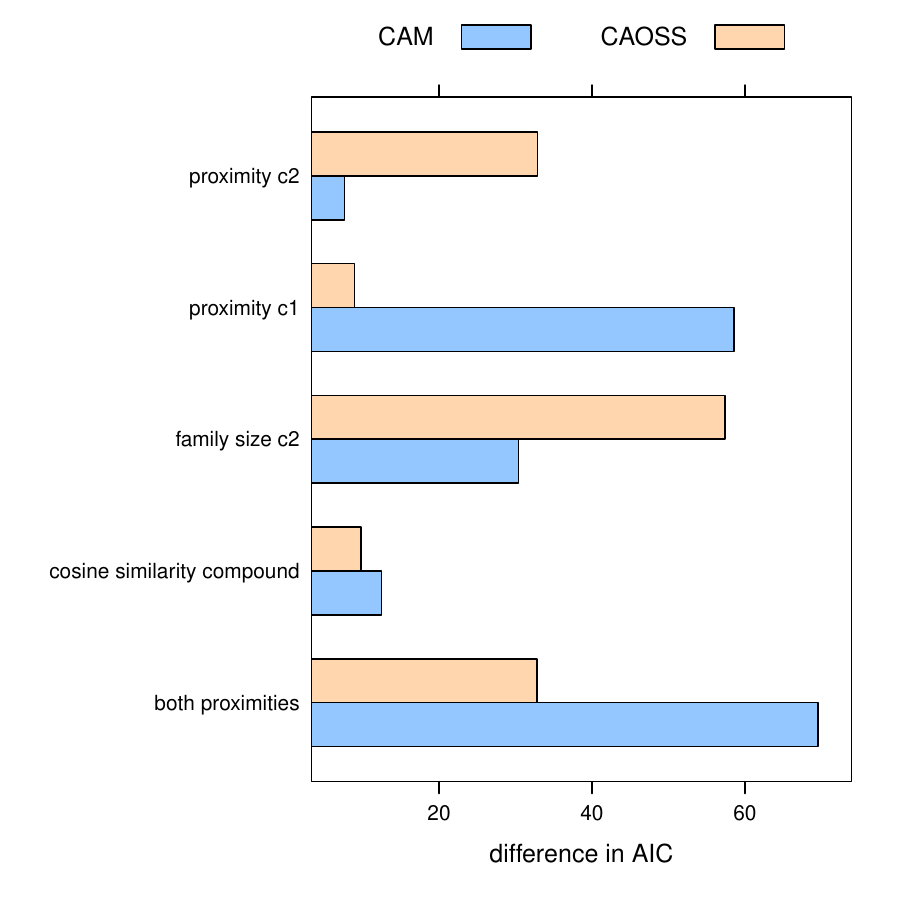}
    \caption{Comparison of variable importances for CAM-based (blue) and CAOSS-based (orange) predictors, as well as second constituent family size, assessed with the increase in AIC when a predictor is withheld from the GAM.}
    \label{fig:aics_barchart}
\end{figure}

Figure~\ref{fig:aics_barchart} facilitates comparison of the relative variable importances of the predictors (leaving aside word frequency, and also predicted compound length, which is relevant only for CAOSS).  The proximity of the second constituent is more important for CAOSS, and that of the first constituent is more important for CAM.  Second constituent family size is more important for CAOSS. The cosine of predicted and observed compound embeddings is slightly more important for the CAM model.

{\color{black}
A final observation is that when the length of the gold-standard compound embedding is brought into the model specification as an additional predictor in the CAOSS and CAM models, the AIC of the CAOSS model reduces from -11,570.10 to -11,672.38, and that of the CAM model from -11,597 to -11,695.46. All other terms in the model remain significant (see the supplementary materials for further details). This suggests that the model-based predictors have explanatory value over and above the L2-norm of the gold-standard compound embeddings.}

\section{General Discussion}

The CAOSS model \citep{gunther2019enter,gunther2020semantic} proposes that the cognitive system predicts the meaning of a compound from the meanings of its constituents, using a linear transformation operating on word embeddings. The present study proposes an alternative model, the Compound Analogy Model (CAM), that is grounded in analogical reasoning over the compound families of the  constituents of compounds.  For each of a compound's constituents, the CAM model starts out with the embedding of that constituent and adds to this the centroid of the shift vectors of that constituent's compound family members. This is done for both constituents, resulting in two --- CAM-based ---`transcended' \citep{libben2010compound,Libben:2014} constituent meanings, which are averaged to obtain the predicted embedding of the compound. 

\textcolor{black}{CAOSS and CAM do not merely differ in their computational implementation; they instantiate different assumptions about the cognitive processes involved in compound interpretation. CAOSS assumes that semantic composition can be captured by a learned transformation operating across compounds, whereas CAM assumes that speakers exploit local regularities emerging from lexical neighborhoods.}

We evaluated the CAOSS and CAM models on a dataset of Mandarin compounds.  We first considered prediction accuracy. For both models, performance was worst for words with two single-syllable constituents, and increased substantially for compounds with three or four single-syllable constituents (henceforth, `characters', as provided by the simplified Mandarin logographic writing system).  The CAM model consistently outperformed CAOSS on training data, and also outperformed the CAOSS model for held-out data with 2 and 4 characters.  It was only for 3-character words for held-out data that CAOSS provided higher accuracy as compared to CAM.  We also examined vector length, using the Euclidean length of the vector, i.e., its L2-norm. The L2 norms of the predicted compound embeddings  were shorter than those of their gold-standard counterparts. The difference in length was substantially larger for the CAOSS model than for the CAM model.  Considered jointly, these computational experiments indicate that the CAM model is a competitive alternative to the CAOSS model. Unlike CAOSS, which requires estimating a $200 \times 400$ linear transformation matrix (80,000 parameters) to map concatenated constituent embeddings into the compound space, \textcolor{black}{CAM is a non-parametric analogy-based model that nevertheless generalizes well to unseen compounds.}

Although for both CAOSS and CAM, accuracy increased for compounds with more than two constituents, upon closer inspection the two models  performed differently for these longer words as compared to the two-character compounds. Constituent families  are extremely small for compounds with three or four characters. As a consequence, the analogical basis for CAM-based generalization is often reduced to a single compound exemplar for each constituent, and analogy collapses. Thus, for the CAM model, predictions for three or four character compounds in the training data are therefore based on the compound itself whenever the compound families of the constituents have only one member. For held-out data, the compound embedding predicted by CAM reduces to just the mean of the two constituent embeddings.  In other words, for longer words, the analogy motivating CAM degenerates. Nevertheless, the model provides predictions that are competitive with those of CAOSS.   

The small constituent families for longer words also affect the performance of CAOSS. Inspection of the constituent-specific transformation matrices of CAOSS shows that for 2-character words, these matrices mainly implement addition of the constituent embeddings: the diagonal elements have large values, whereas off-diagonal elements have very small values. In other words, the two transformation matrices are close to identity matrices.  Interestingly, for 3 and 4 character words, the CAOSS matrices have more larger off-diagonal weights, indicating that for these longer words, CAOSS is picking up some general systematicities and is not just adding up the embeddings of the constituents. As the longer compounds are semantically more transparent, the CAOSS model appears to come into its own for exactly the semantically more regular words.

\citet{hsieh2025compositional} showed that the cosine similarities of the transcended constituent embeddings and the predicted compound embedding, the first and second constituent proximities, are predictive for visual lexical decision latencies to Chinese compounds written with traditional Chinese characters.  We replicated their results with compounds written with simplified Chinese characters, improving on their regression model by allowing lexical predictors to have non-linear effects, using the generalized additive model \citep[GAM,][]{Wood:2017}.  We also documented an effect of the cosine similarity of the predicted and gold standard compound embeddings, in a non-linear interaction with the family size of the second constituent.  Inspection of the geometry of the proximity measures, the effects of which were linear, revealed that they approximate the L2 norm (length) of the predicted compound vector, and that when the proximity measures are replaced by the L2 norm of the predicted embedding, model fit improves. These findings raise questions about what exactly the CAOSS proximity measures tap into. They may gauge not much more than the amount of evidence for lexicality: embeddings with near zero length, which are close to the origin of the semantic space, are not well-learned and not well-distinguishable from nonwords in the visual lexical decision task. The positive correlation of embedding length and word frequency fits well with this explanation.  The only evidence for a semantic transparency effect for visual lexical decisions would then be the small effect of the cosine similarity of the observed and predicted compound vectors. 

A GAM with CAM-based predictors outperformed the GAM with CAOSS-based predictors by more than 20 AIC units. This model included a tensor product smooth for the two proximity measures, and also included the cosine similarity of predicted and observed compound embeddings, again in interaction with the family size of the second constituent. These results indicate that the CAM model is a worthy competitor of the CAOSS model not only as a computational algorithm, but also  when the task is to predict lexical decision times. 

The present findings have several implications for theories of lexical processing.  \textbf{First}, there is no way in which the meanings of two-character compounds can be predicted with sufficient accuracy from the meanings of their constituents.  Accuracy is at 25\% for CAOSS, and better for CAM, but still  only at 50\%.  As a consequence, theories positing obligatory decomposition for reading cannot fall back on a general computational algorithm that will predict the meaning of a compound from the meanings of its decomposed parts.  

Since both CAOSS and CAM operate on embeddings, the  question arises of how in the lexical decision task these embeddings are obtained from the visual input. What is required is some mechanism that takes one or more characters as input, and produces the corresponding embeddings. One such mechanism is provided by a linear mapping that predicts embeddings from character n-grams. We therefore trained such a network on the two-character compounds and their single-character  constituent words in our dataset (see section \ref{sec:compexp}, using the  \textbf{JudiLing} package \citet{Heitmeier:Chuang:Baayen:2026}. With end-state learning (multivariate multiple regression) and form embeddings based on character bi-grams, the network model performs with 100\% accuracy.  In other words, if attention is focused on the first character (requiring decomposition), the network produces the embedding of the first character, and if attention is focused on the two characters jointly (whole-word processing), it will deliver the embedding of the compound.  It turns out that when the L2-norms of the compound embeddings predicted by this network are added as a predictor to the GAM with CAM-based predictors, AIC decreases by some 96 units, supporting the involvement of a direct route from a compound's written form to its embedding. This simple network thus offers a useful approximation of how during reading the embeddings that constitute the input for the CAOSS and CAM models, are obtained during the reading of two-character compounds.

\textbf{Second}, as pointed out by \citet{schafer2020constituent} for English, the higher the uncertainty in analogical reasoning is about the concept represented by the head noun (based on the analogical set of compounds sharing that head noun), the greater the diversity of interpretations is across speakers, and the more these speaker struggle to come up with a meaning for a novel compound. We anticipate that in the CAM model, the quality of the cluster of shift vectors for the head will capture part of this insight. The more widespread in semantic space the shift vectors of compounds sharing a given head are, the more noisy the centroids of these shift embeddings are expected to be, and the more uncertainty there will be concerning what a constituent's prototypical semantic contribution is.   

\textbf{Third}, the CAM model is an analogical model, and thus has the potential of accounting for key findings motivating the CARIN theory of conceptual combination \citep{Gagne:Shoben:97,Gagne:Spalding:Ji:2005}. Consider, for instance, the \textsc{made of} relation in compounds such as \textit{silk shirt, cotton shirt, corduroy shirt, flannel shirt}, and \textit{linen shirt}. The shift vectors from the materials to \textit{shirt} will originate in a cluster of clothing materials in semantic space, and the centroid of the corresponding shift vectors may well capture what it means for a shirt to be made of something.\footnote{However, just as English plural shift vectors vary depending on the semantic class of the noun \citep{shafaei2024semantic}, \textsc{made of} shift vectors are expected to vary consistently depending on the type of object involved. Thus, \textit{wooden chair, metal chair, plastic chair, leather chair} are expected to have a different centroid shift vector than compounds such as \textit{silk shirt, cotton shirt, corduroy shirt, flannel shirt}, and \textit{linen shirt}.}  An important caveat for analogical models addressing the structure and meaning of compounds is that for compounds with more than two constituents, the basis for analogical reasoning can shrink rapidly to just single exemplars.  This is not an issue just  for CAM model, but extends to the CARIN model and to models for compound stress \citep{plag2010compound} and for models predicting the choice of interfixes in German and Dutch compounds \citep{Krott:Baayen:Schreuder:2001,Krott:Libben:Jarema:Dressler:Schreuder:Baayen:2004}.\footnote{Technically, this problem can be attenuated by increasing the set of compounds taken into consideration. For instance, \citet{wang2025composition} trained their implementation of the  NOTCH model on some 500,000 words, obtaining excellent prediction accuracy. However, it is unlikely that individual language users know, or would need, such a huge vocabulary in order to understand compounds. The Modern Chinese Dictionary \citep{modernchinese2016} contains approximately 65,000--70,000 lexical entries that are in common usage in the language community, but the vocabulary of individual speakers, depending on their level of education, is likely to be considerably less.   Our dataset comprises some 33,800 words, which we conjecture to be already overestimating the lexical knowledge of many educated speakers.}

\textbf{Fourth}, the CAOSS model is conceptualized in a general approach to morphology in which each word formation process is associated with its own linear transformation. For derived words, the  FRACCS model \citep{marelli2015affixation} proposes a linear transformation that takes the embedding of the base word as input, and delivers the embedding of the derived word as output. With respect to compounds, the CAOSS model generalizes this approach for the case that there are two `base words'.  However, the FRACCS model has been observed to overfit and to fail to generalize well to test data \citep{stupak2022inquiry}. The present study documents a similar problem for Mandarin compounds.  A further problem is that it is far from clear how the cognitive system would know when to set up the individual transformation matrices. Should English \textit{un-} attached to adjectives, meaning `not', and \textit{un-} attached to verbs, meaning `reverse previous action' be handled by the same transformation? What about comparative \textit{-er} and agent nouns with \textit{-er}?  Ideally, one would want to avoid having to hand-engineer these kind of decisions. The CAM model, on the other hand, straightforwardly generalizes to derivation, and does not require setting up separate mechanisms for individual affixes and compounds. A property of derived words is that the constituent families  of affixes tend to be much larger than those of the constituents of compounds. When an affix is transparent, the centroid of its shift vectors will be well-defined and support generalization. At the same time, the morphological family of the base word, which typically is orders of magnitude smaller, can also be taken into account, just as in the case of compounds.  

Returning to compounding in Mandarin, it is surprising that analogical reasoning can get the meanings right for nearly half of the two-character compounds on which it is `trained'.  Many two-character words are very opaque. Consider, for example, 马虎, {\em ma3hu1}, `careless', has no relation with any constituent meaning of 马 `horse' nor 虎 `tiger'. In the case of 如果 (\textit{ru2gou3}), the first character has meanings such as `be in compliance', `like', `exceed', and `for instance', and the second constituent realizes meanings such as `fruit', `result', `consequence', `as expected'. The meaning of 如果 is simply `if'.  

Nevertheless, in spite of the many two-character words that severely challenge any L2 learner of Chinese, there is considerable analogical structure in the repository of two-character words. This structure is too local to allow robust generalization to unseen two-character compounds. Longer compounds are considerably more transparent than many two-character compounds (e.g., 电话号码, dian4hua4-hao4ma3, `telephone number'), and unsurprisingly, prediction accuracy increases substantially for both CAOSS and CAM.

In summary, we have made a case that compound interpretation is grounded in local analogical generalization.  For familiar compounds, analogical generalisation may exploit  systematicities in the semantic space to support compound interpretation, complementing the compound embeddings as  directly accessed from their written forms.  For novel compounds, the models may provide the gist of the intended meaning. However, as shown by \citet{schafer2020constituent} for English, understanding a novel compound is most likely to heavily depend on the context of use, rather than on de-contextualized constituent embeddings.

\bibliography{references}

\end{document}